\documentclass[twoside,11pt]{article}

\usepackage[left=2.00cm, right=2.00cm, top=2.00cm, bottom=2.00cm]{geometry}
\usepackage[latin1]{inputenc}
\usepackage{amsmath}
\usepackage{amsfonts}
\usepackage{dsfont}
\usepackage[english]{babel}
\usepackage{upgreek }
\usepackage{enumitem}
\usepackage{float}
\usepackage{amsthm}
\usepackage{caption}
\usepackage{algpseudocode}
\usepackage{algorithm}
\usepackage{color}
\usepackage{amssymb}
\usepackage{graphicx}
\usepackage[pagebackref,colorlinks=true,urlcolor=blue]{hyperref}
\usepackage{authblk}

\makeatletter
\newcommand{\labitem}[2]{%
	\def\@itemlabel{\textbf{(#1)}}
	\item
	\def\@currentlabel{#1}\label{#2}}
\makeatother
\newtheorem{theorem}{Theorem}

\newtheorem{mydef}{Definition}
\newtheorem{coro}{Corollary}
\newtheorem{lem}{Lemma}
\newtheorem{rem}{Remark}
\usepackage{ulem}
\newcommand{\proba}{\mathbb{P}}
\newcommand{\expe}{\mathbb{E}}
\newcommand{\expen}{\mathbb{E}_{N_t}}

\newcommand{\pourtout}[2]{\forall~ #1 ~\in~ #2 ,~}

\newcommand{\B}{\beta}
\newcommand{\Bt}{\beta_t}
\newcommand{\Ind}{\mathds{1}}
\renewcommand{\Pr}{\mathbb{P}}
\newcommand{\Rp}{\mathbb{R}_+}
\newcommand{\es}{\epsilon^{+}}
\newcommand{\ei}{\epsilon^{-}}

\newcommand{\Xe}{\upchi_{\epsilon}}
\newcommand{\bXe}{\comp{{\upchi}_{\epsilon}}}

\newcommand{\ad}{\frac{\alpha}{2}}

\newcommand{\cx}{\delta_{1}}
\newcommand{\cv}{\min(1,\ad-\gamma)}
\newcommand{\cva}{\delta_2}
\newcommand{\dt}{\partial_t}
\newcommand{\mub}{\mu_{\beta_t}}
\newcommand{\maxofU}{M}
\newcommand{\comp}[1]{{}^\mathsf{c}#1}
\newcommand{\seqref}[1]{Section \ref{#1}}
\newcommand{\thmref}[1]{Theorem \ref{#1}}
\renewcommand{\eqref}[1]{Equation (\ref{#1})}
\newcommand{\asumref}[1]{\ref{#1}}
\newcommand{\imref}[1]{Figure \ref{#1}}
\newcommand{\corref}[1]{Corollary \ref{#1}}
\DeclareMathOperator{\argmin}{arg\min}
\DeclareMathOperator{\Var}{Var}
\renewcommand{\hat}[1]{\widehat{#1}}
\renewcommand{\tilde}[1]{\widetilde{#1}}
\newcommand*{\BlackBox}{\ensuremath{\blacksquare}}%

\newif\ifnotes\notestrue
\def\hclement#1{} 

\def\hioana#1{}

\title{Convergence rate of a simulated annealing algorithm with noisy observations}

\author[1,2,3]{Clément Bouttier \thanks{clement.bouttier@airbus.com / clement.bouttier@math.univ-toulouse.fr}}
\author[3]{Ioana Gavra \thanks{ioana.gavra@math.univ-toulouse.fr}}
\affil[1]{Airbus Operations SAS, 316 route de Bayonne 31060 Toulouse Cedex 9, France}
\affil[2]{ENAC, 7 avenue Édouard Belin 31055 Toulouse Cedex 4, France}
\affil[3]{Institut de Math\'{e}matiques de Toulouse, Universit\'{e} Toulouse III  118 route de Narbonne 31062 Toulouse Cedex 9, France}

\date{December, 2016}

\begin{document}

\maketitle
\begin{abstract}%
	In this paper we propose a modified version of the simulated annealing algorithm for solving a stochastic global optimization problem. More precisely, we address  the problem of finding a global minimizer of a function with noisy evaluations. We provide a rate of convergence and its optimized parametrization to ensure a minimal number of evaluations for a given accuracy and a confidence level close to 1. This work is completed with a set of  numerical experimentations and assesses the practical performance both on benchmark test cases and on real world examples.
\end{abstract}


\section{Introduction}\label{sec:introduction}
\hclement{The context}
We are interested in an algorithm that solves the stochastic global optimization problem:
\begin{align}\tag{*}\label{PbmGen}
\mbox{Find}\quad  x^{\star} =\underset{x\in E}{\argmin}~\expe_{\omega} (U(x,\omega)),
\end{align} 
where $x$ is a decision variable belonging to some large space $E$, $\omega$  is a random variable and $U$ is the cost, a positive and bounded real valued function. We do not make any assumption on the regularity of $U$. We only expect it to be rapidly evaluable: typically, $U$ is the result of some short numerical simulation. We do not make any distribution assumption for the random inputs $\omega$ themselves but only on the outputs $U$. We assume the code has some robustness property in the sense that, at some point $x$, it is either infinite for all $\omega$  or bounded uniformly in $\omega$. \\
This problem is twofold: we must both estimate and minimize the expectation of the cost. A simple and general approach consists in the minimization of a sample average of Monte Carlo estimators:
$$ \hat{\expe}_{\omega,N} (U(x,\omega)):= \frac{1}{N}\sum_{i=1}^{N}U(x,\omega_i),$$ for any given i.i.d. sample $(\omega_i)_{1\le i\le N}$ of size N distributed according to the distribution $P_\Omega(x)$ of $\omega$ and for any $x\in E$. Such an estimator consistently estimates $\expe_{\omega} (U(x,\omega))$ for any given $x$. Nevertheless, its accuracy is directly linked to $N$ and thus to the computational effort. One can thus wonder if a computationally efficient procedure using that estimator can return a solution to the initial problem given a certain level of accuracy $\epsilon$.

\subsection{Previous works: different types of algorithms}
There were many attempts to solve this stochastic problem across several research communities. We give a brief survey of them in what follows. \\
In the case $E$ is finite and $U$ takes its values in $[0,1]$, problem (\ref{PbmGen}) is labelled as a "simple regret bandit optimization problem" by the bandit community. Indeed it can be seen as the problem of choosing, among a small finite set of slot machines providing random rewards, the one with the best expected reward by playing a minimal number of times. This is not the classical setting of bandit optimization which usually seeks for the "cumulative regret". As algorithm proposals for the simple regret context often extends cumulative regret concepts \cite{audibert2010best}, we focus on them first. The Upper Confidence Bound (UCB, \cite{auer2002finite}) algorithm aimed at building sequences of confidence bounds around the estimated expected cost of each element of the search space. If the space is too large this can be prohibitive. These were several attempts to bypass this issue by adding some assumptions on the regularity of the cost function around its optimum. We can mention HOO \cite{bubeck2011xarmed} that produced guarantees about the cumulative regret for a continuous Lipschitz  cost function  with known Lipschitz constant. In the same framework StoSOO \cite{munos2014bandits} relaxed this last assumption. Both algorithms were however not very efficient in practice if the search space is multidimensional. Indeed they still required some uniform exploration of the state space in the first phase. This could lead to numerical difficulties when the dimension was growing. The Adaptive-treed bandit algorithm \cite{bull2015adaptive} partially solved this issue by adapting the exploration step using a Lipschitz constant per dimension.\\
These algorithms could all be viewed as stochastic variations around the classical \textit{branch and bound} algorithm \cite{little1963algorithm}, which was extensively studied by the optimization community. We can mention the very popular DIRECT algorithm \cite{jones1993lipschitzian}, from which the StoSOO procedure was inspired.\\
Finally, let us mention the computer experiment community that introduced another popular global optimization method for dealing with the stochastic case, the so called Efficient Global Optimization (EGO, \cite{jones1998efficient}) based on expected improvement. The convergence rate of this method was already investigated in \cite{bull2011convergence} in a noise free context. This algorithm focused on minimizing the number of cost evaluations because it considered a setting where the cost evaluations were very time consuming. As a result, in order  to select each evaluation point, it required a higher computational effort and memory storage per iteration than other optimization methods. Such a method could therefore turn  out to under-perform in a setting where the computational cost ratio between selection and evaluation was inverted.\\
A typical algorithm that was known to perform well in the case of time-cheap cost evaluations was the simulated annealing (SA) as mentioned by Locatelli in \cite{horst2013handbook}: \textit{"The latter algorithms} (mainly EGO) \textit{often outperform SA algorithms from the point of view of the number of function evaluations to reach a given relative accuracy on the standard test functions from \cite{dixon1978towards}, but usually require a higher computational effort per iteration. Typical advantages of SA algorithms are their very mild memory requirements and the small computational effort per iteration. If the cost of a function evaluation is very high, then even a considerable computational effort per iteration may be negligible with respect to the cost of a function evaluation, and algorithms which require few function evaluations are preferable; otherwise, also the computational effort per iteration should be taken into account, and from this point of view SA algorithms are often better than other algorithms."}

However SA algorithms have been designed and extensively studied in a context where the exact cost could be observed. We recall below some basic facts in the noiseless case (Section \ref{section SA without noise})  and then present the noisy case which is the setting addressed in this paper (Section \ref{section SA with noise}).

\subsection{Simulated Annealing without noise}\label{section SA without noise}
Let $E$ be some finite search space and $J:E\to\mathbb{R}_+$ a function that we want to minimize, called cost thereafter. 

Simulated Annealing is a classical global optimization method. It aims at building a sequence of elements from $E$ whose last element is drawn from a uniform probability law on the subset of \textit{global} minima of $J$. In other words it aims at sampling from the following distribution 
$$\mu^{\star}= \frac{\mathds{1}_{S_{opt}}}{|S_{opt}|} \mbox{ }, $$
where $S_{opt}=\{x,J(x)=\min_{y\in E} J(y)\}$ and $|.|$ denotes the cardinality of a set. Such a sampling is of course not straightforward but one can notice that this distribution can be rewritten in the following form:
$$ \forall x\in E, \mbox{ } \mu^{\star}(x)=\lim\limits_{T\to 0}\dfrac{e^{\frac{-J(x)}{T}}}{\sum_{y\in E} e^{\frac{-J(y)}{T}}}\mbox{ },$$
and it is well-known that the Gibbs distributions of the form  $\mu_{T}=e^{\frac{-J}{T}}/\sum e^{\frac{-J}{T}} $  are efficiently sampled for reasonably low temperatures  $T\in \mathbb{R}_+$ using the Metropolis-Hastings algorithm \cite{aarts1988simulated}. A quite natural attempt is therefore to build a sequence of sequences obtained using Metropolis-Hastings algorithm for a set of decreasing temperatures. In particular, at a very low temperature, the Metropolis-Hastings algorithm generates exploratory moves that are accepted with very low probabilities, which makes it a very bad sampler.
Therefore it is necessary to first encourage exploration by using a sampling at higher temperatures. A lower bound on the temperature at each step ensuring a convergence in probability of the algorithm has been provided by  \cite{hajek1988cooling}. At the same time another proof of convergence using modern semi-group representation of Markov processes has been obtained by \cite{Holley88Simu}.
The obtained bounds are less explicit but contain information about the convergence rate and the proof scheme is much more general. We set our work in the continuity of this last work and use similar notations. \\

\subsection{Simulated Annealing with noisy evaluations}\label{section SA with noise}
\hclement{the noisy context, review of existing results}
As mentioned previously, our main interest is to extend such a method of simulated annealing to the stochastic case: $$\mbox{Find}\quad  x^{\star} =\underset{x\in E}{\argmin}~\expe_{\omega} (U(x,\omega))$$ where $\omega$ is a random input of a bounded cost function $U$ whose expectation can only be numerically estimated through Monte Carlo simulations. In other words, we consider $J(.)=\expe_{\omega} (U(.,\omega))$. This question is not novel and several attempts were made to address this problem theoretically in the 90's. \cite{Gelfand89NSA} \hclement{Le th\'{e}or\`{e}me est cens\'{e} \^{e}tre repris de l'article \cite{tsitsiklis1989markov} mais il n'y apparait pas explicitement. Les $\Lambda$ dans ce second article sont not\'{e} $R^d$ ou ici $R^\delta$ et il ne s'agit pas de n'importe quel ensemble dans $E$. Ils d\'{e}pendent des $\alpha$. L'application au cas du recuit semble en revanche conforme} were probably the first ones to introduce the notion of simulated annealing with noisy measurements. They assumed an additive Gaussian noise independent of the evaluation point and gave a sufficient condition for the decrease of the variance  $\sigma_k^2$ of this noise, to ensure convergence of the algorithm to the optimal set. \cite{gutjahr1996simulated} extended the results to distributions that are more peaked around zero than the Gaussian distribution. Their convergence result can be stated roughly as follows:\\

\noindent
\begin{theorem}[\cite{gutjahr1996simulated}]\label{thm previous results}
	Let $(X_k)_{k\in\mathbb{N}}$ denote the sequence of states in $E$ visited by the simulated annealing algorithm with Monte Carlo sampling of the noisy measurements. If:
	\begin{enumerate}
		\item[(i)] the convergence conditions from \cite{hajek1988cooling} are satisfied
		\item[(ii)] $\exists \epsilon>0$ such that the standard error of the noise at step $k$ of the algorithm $\sigma_k^2=\mathcal{O}(k^{-(2+\epsilon)})$
	\end{enumerate}
	then $\forall x \in E,\quad \lim\limits_{k\to+\infty}\Pr(X_k=x)=\mu^{\star}(x),$ where $\mu^{\star}$ is the uniform distribution on the global minima of the expected cost. 
	\hfill\BlackBox 
\end{theorem}

This result provided a first answer to our question about the convergence of the algorithm in the stochastic case.
However the convergence statement above did not give any information about the convergence  rate of the algorithm. Following the noise-free proof of \cite{aarts1988simulated}, \cite{homem2000variable} provided an extension of this statement to the noisy case with bounded variance and introducing  a state dependent noise. He obtained the same constraint on the decrease of the variance and the same convergence statement.
He also highlighted the need for an extended result concerning the rate of convergence and for numerical experiments. Indeed on this second point we can mention the works of \cite{fink1998inverse} and \cite{branke2008simulated} that addressed this issue.  \cite{fink1998inverse} made a very interesting proposition in the framework of Gaussian noise.
He proposed to use the noise of measurement to drive the simulated annealing,  \textit{i.e.}, accept a move if the estimated cost of the proposed solution is lower than the one of the current solution. Using an analogy with the Glauber acceptance mechanism, which is a symmetric alternative to the Metropolis-Hasting mechanism \cite{aarts1988simulated}, he proposed a far more efficient criteria for the variance decrease, \textit{i.e.}, $\sigma_k=\mathcal{O}(log(k)^{-2})$. Unfortunately he only provided a few numerical examples to validate his statement and a theoretical proof is still missing.

\subsection{Main contributions}  
In this paper we consider a simulated annealing algorithm based on mini-batches of increasing size. More precisely, at each iteration, the expected cost is estimated by Monte Carlo sampling of increasing sizes. The estimated cost at step $k$ of the algorithm is thus  $\hat{\expe}_{\omega} (U(x_k,\omega))= 1/N_k\sum_{i=1}^{N_k} U(x_k,\omega_i)$, where $N_k$ is an increasing sequence and $\omega_i$ are i.i.d. random variables having the same law as $\omega$. The cost can be written also as $\hat{\expe}_{\omega} (U(x_k,\omega))= \expe_{\omega} (U(x_k,\omega)) +\zeta_\omega(x_k)$, where $\zeta_{\omega}(x_k)$ is some bounded random variable. We denote $\sigma_k^2:=\Var(\zeta_\omega(x_k))$, the variance of post-sampling noise. As it is directly linked to the number of measurements made during the mini-batch, it can be tuned by the user.\\
\paragraph*{Rate of convergence for all variances of polynomial decay.}
In the sequel we first show that theoretical guarantees of Theorem \ref{thm previous results} can be extended to sub-Gaussian random variables (e.g., bounded noise distributions) with stronger convergence results for this algorithm. Indeed we show that convergence can be ensured if the number of measurements is chosen such that $\sigma_k=\mathcal{O}(k^{-(\alpha/2)})$  with $\alpha >0$, which corresponds to $N_k$ of the order $k^{\alpha}$. One can observe that, as opposed to \cite{gutjahr1996simulated}, the convergence still holds for $\alpha\le 2$. This is summarized in \thmref{thm:maintheorem}.\\

We derive the rate of convergence of the procedure (\thmref{thm:conv_speed}) and optimize it (\corref{corollary}) with respect to the noisy simulated annealing algorithm parameters in order to provide a minimal total number of measurements at given accuracy and confidence requirements. This leads to the optimal value $\alpha=2$ for which the number of cost evaluations increases fast enough to ensure almost the same convergence rate as in the noise-free case. This shows that the convergence rate is limited by the concentration speed of the Gibbs measure around its modes. According to our concentration result, increasing  the estimation effort cannot increase the performance of the algorithm above this limit. On the other hand the convergence still holds for a decreased estimation effort ($\alpha<2$) as soon as the cooling schedule is slowed consequently. \hioana{The value of $\alpha$ influences the size of the mini-batches, $N_k$ being of the order $k^{\alpha}$.} \\
\paragraph*{Computational cost in the general case.} Finally, we derive an upper bound on the computational time-complexity of our simulated annealing algorithm (with noisy measurements). This quantity is roughly of the order of:
$$e^{\frac{C_1\log\frac{1}{\delta}}{\epsilon}},$$
where $C_1$ is some constant depending on the cost function itself as detailed in \corref{corollary}. The provided bound exhibit an exponential dependency in $1/\epsilon$ and $\log 1/\delta$. This is comprehensive regarding the generality of the considered problem.\\
\paragraph*{Computational cost in the absence of local minimum}
If the function has no local minimum apart from  the global minimum (e.g., a convex function evaluated on a finite set) the temperature schedule can be adapted and the computational cost becomes of the order of:

$$\left(\frac{C_2\log\frac{1}{\delta}}{\epsilon}\right)^3,$$
where $C_2$ is a constant detailed in \corref{corollary2}. This second bound increases in a polynomial way with respect to $1/\epsilon$ and $\log 1/\delta$. This is a very positive result as it shows that the noisy simulated annealing algorithm recovers the state-of-the-art convergence guaranties if stronger hypotheses on the cost are considered.

\paragraph*{Numerical experiments.} We provide numerical evidence indicating that the numerically observed requirements in \cite{fink1998inverse}, \textit{i.e.}, $\sigma_k=\mathcal{O}(\log(k))$, do not hold for a Metropolis-Hastings Acceptance criteria. We apply the noisy simulated annealing on classical non convex optimization test cases with different level of noise, but also perform a test on a real-world example, \textit{i.e.}, an aircraft trajectory optimization problem using a black-box aircraft performance model.

\subsection{Aircraft trajectory optimization}

As a leading example for this setting, we consider the problem of optimizing commercial aircraft trajectories with respect to a combination of fuel consumption and flight duration:.
\begin{align*}
\text{Find }& u^\star=\underset{u}{\argmin}~ g(x(t_f),t_f) +\int_{t_0}^{t_f}-\dot{m}(x(s),u(s))ds\\
s.t.~& \forall t>t_0 ~ \dot{x}(t)=f(x(t),u(t))\\
& x(t_0)=x_0\\
& d(t_f)=d_f,
\end{align*}
where $x$ is the state of the aircraft, $m$ its mass, $\dot{m}$ its instantaneous fuel consumption, $d$ the ground distance it has flown over, $u$ the path control, $f$ the instantaneous dynamic and $g$ the terminal cost function. The path control $u$ is the combination of the thrust rating $\delta_T$ and the lift coefficient $C_l$.
\begin{align*}
u&=\left(\begin{array}{c}
C_l\\
\delta_T
\end{array}\right)
\end{align*}
Estimates of the cost of trajectories are usually obtained through numerical integration of the flight dynamic equations, $f$:
\begin{align*}
\dot{x}=\left(\begin{array}{c}
\dot{V}\\
\dot{\gamma}\\
\dot{h}\\
\dot{d}\\
\dot{m}	
\end{array}\right)&=\left(\begin{array}{c}
(T(h,V,\delta_T)-D(h,V,C_L))\frac{1}{m}-g\sin \gamma\\
(L(h,V,C_L)-mg\cos \gamma)\frac{1}{mV}\\
V\sin(\gamma)\\
V\cos (\gamma)\\
\eta T(h,V)
\end{array}\right)
\end{align*}
where $T$ is the thrust, $D$ the drag, $L$ the lift, $\eta$ the specific fuel consumption, $\gamma$ the path angle, $V$ the speed of the aircraft and $h$ its height. These equations involve some terms like the aerodynamic drag coefficient ($C_D$) or maximal propulsion effort ($T_{max}$)  who are estimated using interpolation of experimental local measurements. 
\begin{align*}
T(h,V,\delta_T)&=\delta_TT_{max}(h,V)\\
L(h,V,C_L)&=\frac{1}{2}\rho(h,V)SV^2C_L\\
D(h,V,C_L)&=\frac{1}{2}\rho(h,V)SV^2C_D(C_L,V)
\end{align*}
No analytic solution is therefore available nor conceivable. Moreover the relation between cost and trajectory control parameters cannot reasonably be assumed to be convex.
At last, the cost estimation relies on some predicted flight conditions including atmospheric ones. Hence, real-flight costs can thus deviate substantially from their predictions and some uncertainty propagation method must be applied to obtain an accurate estimate of the expected flight costs. In other words the function we want to minimize can only be evaluated with a certain random error, which corresponds exactly to the setting of this paper. Finally, the computational efficiency is a key ingredient as it must be performed only a few hours before the planned flight. For more information about aircraft trajectory optimization we refer to \cite{Betts1998}.\\
This example completely fits our requirement as the computation of the cost of one single complete trajectory is quite fast, \textit{i.e.}, less than a second. Therefore, the EGO  algorithm \cite{jones1998efficient}  would not be suited for this application.
On the other hand formulations based on the DIRECT algorithm \cite{jones1993lipschitzian} would suffer strongly from the dimension of the problem. An additional element that motivates the use of simulated annealing is the fact that in the case of trajectory optimization the set of admissible controls is not known in advance as it is path dependent.
We can only ensure that this set is connected. This implies in particular that no projection on the constraints can be performed and excludes the projected stochastic gradient descent for example. In the case of simulated annealing, a very simple step can bypass this issue.
By setting the value of the cost to infinity when the trajectory evaluator returns an error we ensure staying in the admissible domain. Consequently, a feasible solution and a conservative approximation of the admissible domain are the only requirements to initiate the algorithm in this setting. 

\subsection{Outline of the paper}
Our paper is organized as follows. In \seqref{sec:convergence_statement} we present the noisy simulated algorithm and our main theoretical result. In Sections \ref{sec:inf_gen}, \ref{sec:gen_diff} and \ref{sec:gronwall} we provide the proof of this statement. More precisely, in \seqref{sec:inf_gen} we compute the infinitesimal generator of the noisy simulated annealing algorithm. In \seqref{sec:gen_diff} we compare it to the one of the noise-free simulated annealing algorithm from \cite{Holley88Simu}. This enables us to derive a differential inequality for a $L^2$ distance between the distributions of the two previously mentioned processes. Integrating by applying Gr\"{o}nwall's Lemma  \seqref{sec:gronwall}, we obtain obtain our convergence result. In the same section, we show how to tune the parameters of the algorithm in order to optimize the performance bound and give the corresponding computational cost. In \seqref{sec:num_res} we propose some numerical insight on synthetic and real data experiments. 
\section*{Acknowledgments.}
We thank S\'{e}bastien Gadat for introducing this topic to us, making this collaboration possible and for fruitful discussions and helpful insights, and S\'{e}bastien Gerchinovitz for all his constructive advice and ideas. 

\section{Noisy Simulated Annealing algorithm: statement and convergence result} \label{sec:convergence_statement}
We first present our extended version of the simulated annealing to the stochastic case, whose pseudo-code can be found in Algorithm \ref{NSA}.
\subsection{Noisy Simulated Annealing algorithm (NSA)}
\begin{algorithm}
	\caption{Noisy Simulated Annealing}
	\label{NSA}
	\begin{algorithmic}
		\Procedure{NSA}{\textbf{Inputs:} Neighbourhoods structure $(S_x)_{x\in S}$, Initial guess $x_{0}$,
			increasing function $\beta:\mathbb{R}_+\to\mathbb{R}_+$, Function $t\mapsto n_t$}\\

		\State Initialize time $t_0=0$
		\State $\beta_0=\beta(t_0)$\\
		
		\For{k from 0 to Maximal number of iterations}

		\State Draw  one solution candidate: $\tilde{x}_{t_k}\in S_{x_{t_k}}$ according to $q_0(x_k,\cdot)$
		\State Draw $N_{t_k}\sim \mathcal{P}oisson(n_{t_k})+1$
		\State Draw $2N_{t_{k}}$ simulation conditions independently:
		\State $ \quad(\omega_1^k,...,\omega_{N_{t_k}}^k)\sim \left(P_\Omega(x_{t_k})\right)^{\otimes N_{t_k}}$ and $(\tilde{\omega}_1^k,...,\tilde{\omega}_{N_{t_k}}^k)\sim\left(P_\Omega(\tilde{x}_{t_k})\right)^{\otimes N_{t_k}}$
		\State Compute estimates $\hat{J}(x_{t})$ and $\hat{J}(\tilde{x}_{t})$ using the $N_{t_k}$  conditions:
		\State$\quad\hat{J}(x_{t_k})=\frac{1}{N_{t_k}}\sum_{i=1}^{N_{t_k}}U(x_{t_k},\omega_i^k)$,
		\State $\quad\hat{J}(\tilde{x}_{t_k})=\frac{1}{N_{t_k}}\sum_{i=1}^{N_{t_k}}U(\tilde{x}_{t_k},\tilde{\omega}_i^k)$\\
		\State Draw an exponential random variable $\xi_{k+1}$ of parameter $1$
		\State Update time $t_{k+1}:=t_k+\xi_{k+1}$\\
		\State With probability $e^{-\beta_k\lfloor \hat{J}(\tilde{x}_{t_k})-\hat{J}(x_{t_k})\rfloor_+}$:
		\State $\quad$ set $x_{t_{k+1}}:=\tilde{x}_{t_k}$ 
		\State Otherwise set $x_{t_{k+1}}:=x_{t_k}$\\ 
		\State Increase the inverse of the temperature $\beta_{k+1}:=\beta(t_{k+1})$	
		
		\EndFor
		
		\State \textbf{return} $x_{t_{k+1}}$
		\EndProcedure
	\end{algorithmic}
\end{algorithm}
where $\lfloor x\rfloor_{+}=0$ if $x\le 0$ and $x$ if not. \\
As in the deterministic setting, the algorithm requires an initial feasible solution $x_{t_0}$, a temperature schedule $T_t$ (we will mostly use its inverse $\beta_t=1/T_t$), and a good neighbourhood structure. What we mean by good will be specified in the definition of  \asumref{asum:neighboor}. The algorithm explores the state space in the following manner. After $k$ iterations, at time $t_k$, it selects a random neighbouring solution $\tilde{x}_{t_k}\in S_{x_{t_k}}$ ($S_{x_{t}}$ being the set of neighbours of $x_{t_k}$) according to a proposition law. Then it compares the estimate $\hat{J}(\tilde{x}_{t_k})$ of the cost of this new solution to the estimated cost $\hat{J}(x_{t_k})$ of the current solution and then it decides to substitute (or not) the new to the current:\begin{itemize}
	\item if the estimated cost of the new state is lower than the current one, \textit{i.e.}, $\hat{J}(\tilde{x}_{t_k})\le\hat{J}(x_{t_k})$, the move is accepted, \textit{i.e.}, $x_{t_{k+1}}\leftarrow \tilde{x}_{t_k}$
	\item if not, it is only accepted with a probability $\exp(-\beta_{t_k}( \hat{J}(\tilde{x}_{t_k})-\hat{J}(x_{t_k}))$.
\end{itemize}  The time $t$ is then updated using independent exponential random variables, enabling us to consider the NSA as a continuous time Markov process.	
\subsection{General setting and notations}\label{assumption_subsec}
To state the convergence of Algorithm \ref{NSA}, we first need to describe formally the framework we are working in.
Notations introduced in this section are valid for the whole paper unless mentioned explicitly.
\begin{itemize}
	\item Regarding the noise structure and the estimation procedure, we denote:
	\begin{itemize}
		\labitem{$\hat{J}$}{asum:noise} the estimated cost:  $\hat{J}:E\to\mathbb{R}_+$, such that $\forall~x\in E,\hat{J}(x)=\frac{1}{N}\sum_{i=1}^{N}U(x,\omega_i),$
		where: $(\omega_1,...,\omega_{N})$ is a $N$ i.i.d. vectors sequence drawn from distribution $P_\Omega(x)$
		\labitem{$\xi_k$}{asum:time_incr} the time increments: $(\xi_k)_{k\in \mathbb{N}}$ is a sequence of i.i.d. exponential random variables of parameter $1$ 
		\labitem{$t_k$}{asum:time} the jumping times: $\forall k \in~\mathbb{N},\quad t_k=\sum_{i=1}^{k}\xi_i$.
		\labitem{$n_t$}{asum:poisson_mean} the samplig intensity: $n_t$  a continuous increasing function.			
		\labitem{$N_{t_k}$}{asum:poisson} the sample sizes: $N_{t_1},N_{t_2}, \dots N_{t_n}$  are independent  for all $ n\in \mathbb{N}$ and all $ 0<t_1<t_2\ldots<t_n$  and  $$N_{t_k} \sim \mathcal{P}oisson(n_{t_k})+1,$$
	\end{itemize}
	We can make a few remarks about the different notations. The construction of $N_t$ ensures that its value is a strictly positive integer at all times. The reason why we choose to have a randomly sized sample for the Monte Carlo estimation procedure is rather technical. It enables generating a continuous transition probability as it can be noticed in \eqref{eq:def_qtilde} and ease the formulation of the infinitesimal generator (\eqref{nsa generator}).	
	\item About the state space, we denote:
	\begin{itemize}
		\labitem{$E$}{asum:Efinite} a finite state space.
	\end{itemize}
	\begin{itemize}
		\labitem{$S$}{asum:Connection} a neighbourhood structure such that $E$ is connected  with respect to it, \textit{i.e.}, $S$ is a connected graph containing all the points in $E$. For any $x$ in $E$, we denote $S_x$ the set of its direct neighbours.  
	\end{itemize}
	\begin{itemize}
		\labitem{$\mu_0$}{asum:initdistrib} the initial distribution, a probability measure that charges every point of a subset of interest $E'\subset E$ defined more precisely in (\ref{asum:Ubounded}) ,
	\end{itemize}
	\begin{itemize}
		\labitem{$q_0$}{asum:neighboor} the proposition law, an irreducible and $\mu_0-reversible$ transition probability, \textit{i.e.}, $\forall x,y \in E,$ $\sum\limits_{n=0}^{\infty} q_0^{(n)}(x,y)=\infty$ and $\mu_0(x)q_0(x,y)=\mu_0(y)q_0(y,x)$. In addition we assume that for any $x$ in $E$, we have $q_0(x,S_x)=1$ 
	\end{itemize} 
	
	Considering a finite search space \asumref{asum:Efinite} enables us to easily derive the spectral gap inequality in \thmref{thHolley} and overcome differentiation-under-the-integral-sign issues in \eqref{lemma6Holley}. Nevertheless, it could be replaced by coercivity assumptions on the function $J$, which could be more general but not really well suited for the application we are looking for. It is our most restrictive assumption. Nevertheless it is in line with previous works on noisy global optimization for example: \cite{gutjahr1996simulated}, \cite{homem2000variable} or \cite{fink1998inverse}. It corresponds to a historical use of simulated annealing for problems with huge finite search space like for the traveling salesman problem \cite{aarts1989boltzmann}. Mimicking \cite{Holley88Simu}, we might however relax this assumption of finiteness. Nevertheless it requires more technicalities as in \cite{aarts1988simulated} and this is left for future work.
	 
	 We assume that the algorithm can visit and start from every point in the solution space through the connection  assumption \asumref{asum:Connection} and the definition of \asumref{asum:initdistrib}. The proposition law \asumref{asum:neighboor} defines the way a new solution $\tilde{x}$ is proposed to the NSA at each iteration. The irreducibility of $q_0$ implies the fact that one can go from any state $x$ to any other state $y$ using the neighbourhood structure $S$, in a finite number of steps.  The $\mu_0-reversibility$ is used to simplify the notations. A classical choice \cite{aarts1988simulated} for $q_0$ and $\mu_0$ is: $\pourtout{x,y}{E} \mu_0(x)=\frac{1}{|E|}$ and $q_0(x,y)=\frac{1}{|S_x|}$, assuming every point in $E$ to have the same number of neighbors. However there are other possible choices for $\mu_0 $ and $q_0$. This last two assumptions are inherited from the classical Metropolis-Hasting sampling algorithm which corresponds to the NSA algorithm with no cooling mechanism and no noise. They ensure that a run in this setting, starting from any point of the search space, converges to a stationary distribution which is the Gibbs measure associated to $J$.
	\item About the cost function, we consider:
	\begin{itemize}	
		\labitem{$U$}{asum:Ubounded} the underlying cost: $\exists M>0$ and $\exists E' \subset E$, such that $U$ is bounded and non-negative on $E'$, i.e. $\forall x~\in~E', \forall \omega$, $0\le U(x,\omega)\le \maxofU$  and $U$ is infinite on $E\backslash E'$, \textit{i.e.},
		$\forall x\in E\backslash E'$,  $\forall \omega$, $U(x,\omega)=+\infty.$
	\end{itemize}
	
	The assumption about \asumref{asum:Ubounded} being bounded is not restrictive. It reflects the practical setting where a simulation code crashes out of the definition domains. We associate infinite costs to crashes and thus (\asumref{asum:Ubounded}) is rather a consequence of (\asumref{asum:Efinite}). 
	
	\item About the algorithm parametrization, we denote
	\begin{itemize}	
		\labitem{$\beta_t$}{asum:tempschedule} the inverse of the temperature: a positive increasing real function of $t$,
		\labitem{$\alpha$}{asum:ntdef} the sampling size (expected number of simulations): $\exists \alpha \in \mathbb{R}_{+}$ such that $n_t=(t+1)^{\alpha},$
	\end{itemize}
	\asumref{asum:tempschedule} is usually chosen such that $\forall t~\in~\Rp$, $\frac{d \Bt}{dt}=\frac{bd}{1+td}$ for some $b,d\in \Rp$, as it was shown by \cite{hajek1988cooling} and \cite{Holley88Simu} to be a necessary condition to ensure the convergence of the simulated annealing algorithm for any cost function. There is no reason to expect that the noisy context would be more favorable than the deterministic one. As suggested by the definition of \asumref{asum:ntdef}, we choose a polynomial growth of the number of simulations for the cost estimation. We show later on  in this work that this ensures the convergence of the noisy simulated annealing for a good choice of $\alpha$ and $b$.	
	
\end{itemize}
\subsection{Tool for the analysis: the NSA process}\label{notations}

We now present the mathematical formalization of the NSA algorithm's underlying stochastic process. First, for pedagogical purposes, we omit the temperature evolution and noisy measurements. The NSA algorithm then becomes a simpler Markov chain exploring the state space $E$ according to the Markovian transition matrix whose elements are of the form:
\begin{align}
\Pr(x\to y)=q_{\beta}(x,y) & =\begin{cases} q_0(x,y) e^{-\beta \lfloor J(y)-J(x)\rfloor_{+}}~ \quad \mbox{if } y\neq x \\
1-\sum\limits_{z\in E\backslash{x}}q_{\beta}(x,z) \quad \mbox{if }y=x, \end{cases}\label{eq:q_intro}
\end{align}
This reflects the transition mechanism introduced at the beginning of this section. As the process is in fact a continuous one, we must also consider the time component. NSA jumps happen at stochastic times and the probability of acceptance depends on these times. Combining the law of the jumping times and the previous mechanism, we can make their joint transition probability explicit: 
\begin{itemize}
	
	\item Let $(\tilde{\upchi}_k,T_k)_{k\in \mathbb{N}} $ a $E\times\mathbb{R}_+$-valued Markov chain such that $\forall k \in \mathbb{N},$ $\forall y \in E$, $\forall u \in \mathbb{R}_+$:
	\begin{align}	 	\Pr(\tilde{\upchi}_{k+1}=y,T_{k+1}\ge u|\tilde{\upchi}_k, T_k)=		
	\int\limits_{u}^{+\infty}\tilde{q}_{\beta_{\tau}}(\tilde{\upchi}_k,y)\Ind_{[T_k,+\infty[}(\tau)e^{-(\tau-T_k)}d\tau,
	\end{align}
	where 
	\begin{align}
	\tilde{q}_{\Bt}(x,y)=\begin{cases}q_0(x,y) \mathbb{E}_{N_t}\mathbb{E}_{\omega_1,...,\omega_{N_t}}\left(e^{-\frac{\Bt}{N_t} \lfloor\sum_{i=1}^{N_t}  U(y,\omega_i)-U(x,\omega_i)\rfloor_{+}}\right) \quad \mbox{if } y\neq x \\
	1-\sum\limits_{x\neq z}\tilde{q}_{\B_t}(x,z) \quad \mbox{if }y=x \end{cases}\label{eq:def_qtilde}
	\end{align} 
	This is a similar construction to the one of the classical simulated annealing process \cite{hajek1988cooling}. The state transition mechanism must also reflect the estimation procedure, therefore the form of \eqref{eq:def_qtilde} differs from \eqref{eq:q_intro}.
	As mentioned before the function $t\mapsto\beta_t$ represents the inverse of the temperature schedule and $N_t$ is the random  process described by \asumref{asum:poisson}. The jumping times, or evaluation times of the process happen at times defined by the sequence $T_k$ (\textit{cf.} the definition of \asumref{asum:time}). 
	
	The chain $(\tilde{\upchi}_k)_{k\ge 0}$ explores the state space $E$ using a transition probability $\tilde{q}_{\beta_t}$ constructed in a same way as the classical one, replacing the exact value of $-\B_{T_k}(J(y)-J(x))_+$ by its Monte Carlo estimation. The expected value from the formula comes from the fact that, as mentioned in the definition of \asumref{asum:noise} and  in Algorithm \ref{NSA}, we use a random number of Monte Carlo shootings for the estimations.
\end{itemize}
Finally, we obtain the NSA process by associating the two sub-processes as follows:
\begin{itemize}
	
	\item Let $\left(\tilde{X}_t\right)_{t\ge0}$ be the inhomogeneous Markov Process such that $\tilde{X}_t= \tilde{\upchi}_k$ if $T_k\le t <T_{k+1}$. One can see that this process  is piecewise constant and jumps at exponential times from one candidate solution to another, in other words  $(\tilde{X}_t)_{t\ge 0}$ is just the continuous-time version of the noisy simulated annealing discrete time process,$(\tilde{\upchi}_k)_{k\ge 0}$.\\
	
	Note that, if $y\in E\backslash E'$ then $\forall x \in E',\quad \tilde{q}_{\beta_t}(x,y)=0$. Hence, if the initial solution $\tilde{X}_0$ is chosen in $E'$, then $\forall t\ge 0,~ \tilde{X}_t\in E'$. 
\end{itemize}
\subsection{Convergence result}
We denote:
\begin{itemize}
	\item $m^{\star}$ the maximum depth of a well not containing a fixed global minimum of the function $J$. To be more precise, we call a path from $x$ to $y$ any finite sequence $x_0=x,x_1,\dots,x_n=y$  such that for all $i$, $x_{i+1}\in S_{x_i}$. Let $P_{xy}$ be the set of paths from $x$ to $y$.\\
	For a given path $p\in P_{x,y}$, the elevation of the function $J$ on $p$ is $ \underset{z\in p}{\max} ~ J(z) $.  Minimizing this quantity over the set of possible paths $P_{x,y}$, gives us the elevation of the cheapest path going from $x$ to $y$. Denote this elevation by: 
	$$H_{x,y}= \underset{p \in P_{xy}}{\min} \left\{\underset{z\in p}{\max} ~ J(z)\right\}$$
	 Then
	\begin{equation}\label{definition of m}
	m^{\star}:=\underset{x,y \in E}{\max}\left\{H_{x,y}- \max\left(J(y),J(x)\right)\right\}
	\end{equation}

	\begin{figure} 
		\centering
		\includegraphics{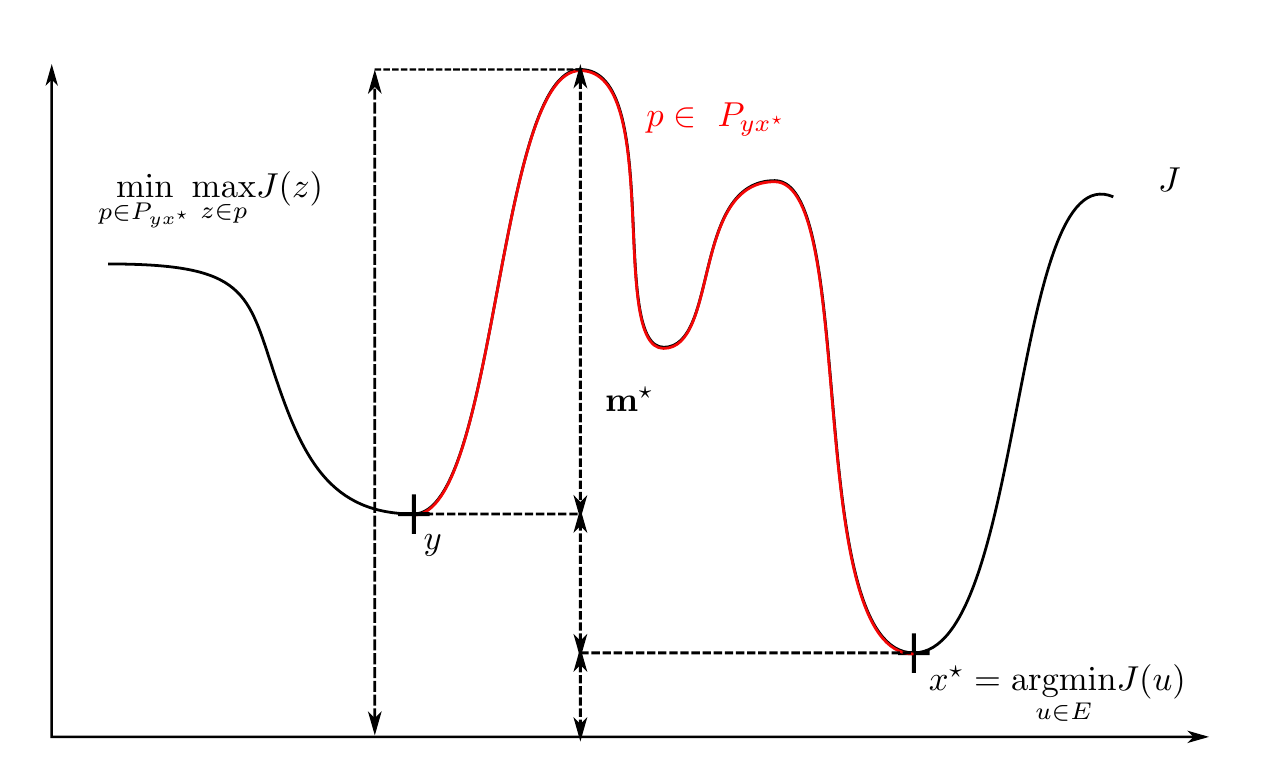}
		\caption{$m^{\star}$, maximal depth of local \hioana{but not global} minima }
		\label{mstar}
	\end{figure}
	As represented on \imref{mstar}, $m^{\star}$ can also be understood as the highest energy barrier to climb to go from one point to another in the search space in the easiest direction. 
	As mentioned before, it also represents the maximal depth of a well not containing a fixed global minimum. If $x^{\star}$ is a global minimum then:
	$$m^{\star}= \max_{y\in E}\left\{ H_{x^{\star},y}-J(y)\right\}.$$ 
	The definition provided here is equivalent to the classical one, \textit{i.e.}, the one provided in \cite{hajek1988cooling} and \cite{Holley88Simu}. A proof of this statement can be found in Appendix \ref{mstardefequiv}. 
	 
	\item  $\gamma(\beta)$ the spectral gap between 0 and the rest of the $L^2(\mu_\beta)$ spectrum of $-L_{\beta}$, where $L_{\beta}$ is the generator of the classical simulated annealing (for more details about $L_{\beta}$ see Section \ref{sec:inf_gen}):

	\begin{align}
	\gamma(\beta):=\inf\left\{-\int \phi L_{\beta} \phi d\mu_{\beta}~ \mbox{s.t. } \int|\phi|^2 d\mu_{\beta}=1 \mbox{ and } \int\phi d\mu_{\beta}=0 \right\}
	\end{align}
\end{itemize}	
Following \cite{Holley88Simu}, we know that given $E$, $\mu_0$ and $U$, there exists a constant $c$ such that:
$$\forall \beta\ge 0,\quad \gamma(\beta)\ge ce^{-m^{\star}\beta}$$
Remark that this lower bound is mainly informative for small values of $\beta$.	
In addition set:  $$\forall x \in~E,~ J(x)=\mathbb{E}_{\omega}(U(x,\omega)) \mbox{ and }J^{\star}=\underset{x\in E}{\min}~\mathbb{E}_{\omega}(U(x,\omega)).$$
We define $\Xe$ the set of $\epsilon$-optimal points in $E$, \textit{i.e.}, 
\begin{align}\label{def:optimality_set}
\Xe=\{ x:J(x)\le J^{\star}+\epsilon\},
\end{align}
and denote $\comp{\Xe}=E\backslash \Xe$, its complementary in $E$. We also write $a\wedge b= \min{a,b}$. \\

\noindent
\begin{theorem}\label{thm:maintheorem}Consider the settings of \seqref{assumption_subsec},\\ if $\beta_t=b\log(td+1)$ and	 $n_t=(t+1)^{\alpha},$ with: $$\{ m^{\star}b<1\wedge\alpha/2\} \mbox{ or } \{ m^{\star}b=1, \ \alpha>2 \mbox{ and } d<2cm^{\star}/M \},$$ then there exits $ C>0$ such that, $\forall t\in \mathbb{R}_+,~\forall \epsilon>0,~\quad  \Pr(\tilde{X}_t\in \comp{\Xe})\le C(\mu_{\beta_t}(\comp{\Xe}))^{1/2}.$  
	\hfill\BlackBox 
\end{theorem}

This theorem is a natural extension of the result provided by  \cite{Holley88Simu}. There are  two main interesting facts to point out. First, we obtain a balance between the expected number of Monte Carlo simulations at each step of the algorithm and the inverse of the temperature, \textit{i.e.}, $\{ m^{\star}b<1\wedge\alpha/2\} \mbox{ or } \{ m^{\star}b=1, \ \alpha>2 \mbox{ and } d<2cm^{\star}/M \}$. Reducing the growth rate $\alpha$ of the number of simulations below the quadratic rate should be compensated by decreasing accordingly the temperature factor $b$.  Second, the convergence is stated in terms of a bound on the probability of not returning  an optimal solution. Using the concentration speed of the Gibbs measure one can deduce a rate of convergence of the algorithm. Also the theorem provides an insight on how the algorithm could be used in practice. A run of parallel noisy simulated annealing would have a probability of returning a bad solution that would decrease in the power of the number of runs. Nevertheless this benefit should be traded with an additional selection cost. Indeed, if we obtain $K$ solutions retrieved by $K$ parallel NSA realization, we still face the problem of selecting the best one. We only access estimates of the costs associated to each solution.\\

\paragraph*{Sketch of the proof}
The proof of this theorem is divided into three parts. First, in \seqref{sec:inf_gen}, we  compute the infinitesimal generator of the classical (\eqref{classic generator}) and noisy simulated annealing (\eqref{nsa generator}). Second, in \seqref{sec:gen_diff}, we compare them (Lemma \ref{bounds}) and third, in \seqref{sec:gronwall}, we conclude about the convergence using the Gr\"onwall lemma (\eqref{gronineq}) and the convergence of the classical simulated annealing (\eqref{concentration}). 

\paragraph*{Convergence rate}
In the case $m^{\star}b<1$ a finer bound can be deduced from Gr\"onwall's lemma and one can obtain a more precise convergence rate for the algorithm (\thmref{thm:conv_speed}), which is roughly of the order of:
	\begin{align*}
	\Pr(\tilde{X}_t\in \bXe) &\le  \Gamma~ t^{((m^{\star}-\epsilon)b-\min(1,\alpha/2))/2}
	\end{align*} 
where $\Gamma$ is some constant detailed in \thmref{thm:conv_speed}.
In particular this implies that for fixed $\epsilon,\delta >0$ we can find $T^{\star}$ such that $\Pr(\tilde{X}_T^{\star}\in \comp{\Xe})\le \delta$. This leads to a bound (Lemma \ref{lemma_tstar}) on the computational complexity, $\mathbb{E}\left(N_{call}^{T*}\right)$, of the order of:
$$\mathbb{E}\left(N_{call}^{T*}\right)\le  \left(\frac{\Gamma}{\delta}\right)^{2(\alpha+1)/(\min(1,\alpha/2)-(m^{\star}-\epsilon)b)}.$$ 

\section{Proof, Part 1: Infinitesimal generator}\label{sec:inf_gen}
In this section we use the semi-group characterization of the generator in order to prove that as soon as $\tilde{q}_{\beta_t}$ defined in \eqref{eq:def_qtilde} is continuous with respect to $t$ then the infinitesimal generator $\tilde{L}_{\beta_t}$ of the Markov process $\tilde{X}_t$ can be written as: 
\begin{equation}\label{nsa generator}
\tilde{L}_{\B_t}f(x)=\sum\limits_{y\in E}\biggl(f(y)-f(x)\biggr)\tilde{q}_{\beta_t}(x,y).
\end{equation}

We briefly recall the definition of the semi-group associated to a Markov process. 
\begin{mydef}
	The semi-group $(P_{t,t+s})_{t\ge 0,s\ge 0} $ associated to the Markov process $(X_t)_{t\ge 0}$ is a family of probability kernels such that for all non-negative borelian functions:
	$$\forall t,s\in \Rp \quad P_{t,t+s}f(x) =\expe(f(X_{t+s})|X_t=x) $$ 
\end{mydef}
Let $(P_{t,t+s})_{t\ge 0,s\ge 0} $ be the semi-group associated to the Markov process $(X_t)_{t\ge 0}$. The semi-group characterization of its generator is given in the following definition:
\begin{mydef}\label{defgenerator}
	The  infinitesimal generator $L_t$ of the Markov process $(X_t)_{t\ge 0}$ is defined as the operator such that for any bounded function $f$:
	$$	L_{t}f(x) = \lim\limits_{s\to 0} \frac{P_{t,t+s}f(x)-P_{t,t}f(x)}{s}$$
\end{mydef}
We start by computing the infinitesimal generator $L_{\B_t}$ of the process associated to the SA algorithm, \textit{i.e.}, with no measurement noise, and then deduce the infinitesimal generator of the NSA algorithm. Using similar notations to the ones of \seqref{notations}, we consider the noise free inhomogeneous Markov process, $(X_t)_{t\ge 0}$ constructed from the inhomogeneous Markov chain $(\upchi_k)_{k\in\mathbb{N}}$ whose  one step transition probability is:
$$\forall x,y \in E,\quad q_{\B_{T_k}}(x,y)=\begin{cases} q_0(x,y)e^{-\B_{T_k}(J(y)-J(x))_+} \quad \mbox{if } y\neq x \\
1-\sum\limits_{z\in E\backslash \{x\}}q_{\B_{T_k}}(x,z) \quad \mbox{if y=x} \end{cases}$$ 

This is the natural extension of the simulated annealing process with discrete jumping times \cite{hajek1988cooling} to the continuous time process. In this configuration, the jumping times are drawn from an i.i.d. sequence of exponential random variables of parameter $1$. In the homogeneous configuration, \textit{i.e.}, $\beta_t=\beta$, the infinitesimal generator has a classical form: $L_{\beta}=Q_{\beta}- Id$ where $Q_{\beta}$ is the transition matrix associated to $q_{\beta}$ and $Id$ denotes the identity. The extension to the generator of the non-homogeneous process is not straightforward. Therefore we propose to detail the computations.

By  definition, for any bounded function $f$:  
\begin{align}
L_{\B_t}f(x) &= \lim\limits_{s\to 0} \frac{P_{t,t+s}f(x)-P_{t,t}f(x)}{s}\nonumber\\ \nonumber
&= \lim\limits_{s\to 0} \frac{\sum\limits_{y\in E}f(y)\Pr(X_{t+s}=y|X_{t}=x)-f(x)}{s}\\
&= \lim\limits_{s\to 0} \frac{\sum\limits_{y\in E}f(y)\Pr(X_{t+s}=y,H_{t+s}-H_{t}\ge 0|X_{t}=x)-f(x)}{s},\\\nonumber
\end{align}
where $H_t=\max\{k \in \mathbb{N}:T_k<t \}$ denotes the number of jumps before time t. Since $T_k$ is a sum of independent exponential variables of parameter $1$, one can remark that $H_t$ is in fact a Poisson process of parameter $1$.
\medskip

In order to compute the above limit, we begin by calculating a more explicit form of the probabilities above. We can divide these computations into three parts according to the number of jumps between $t $ and  $t+s$: 
\begin{align*}
\Pr(X_{t+s}=y,H_{t+s}-H_{t}\ge 0|X_{t}=x)=~&\Pr(X_{t+s}=y,H_{t+s}-H_{t}= 0|X_{t}=x)\\
&+\Pr(X_{t+s}=y,H_{t+s}-H_{t}= 1|X_{t}=x)\\
&+\Pr(X_{t+s}=y,H_{t+s}-H_{t}\ge 2|X_{t}=x).\\
\end{align*}
The first case is straightforward, if there is no jump between $t$ and $t+s$, the process will not change its position and we thus have:
\begin{align*}
\Pr(X_{t+s}=y,~H_{t+s}-H_{t}= 0|X_{t}=x)&= \delta_{x}(y)e^{-s}.\\
\end{align*}
The second case is slightly more involved. Using the stationarity and the definition of Poisson processes, the event that the algorithm goes from $x$ to $y$, having only one jump between $t$ and $t+s$,  can be written as:
\begin{align*}
\Pr(X_{t+s}=y,H_{t+s}-H_{t}= 1|X_{t}=x)
=& \Pr(X_{t+s}=y,\xi'_1<s, s-\xi'_1<\xi'_2 |X_t=x )
\end{align*}
where $\xi'_1$ and $\xi'_2$  are two independent exponential random variables of parameter one. \\
Let $\xi=(\xi_1',\xi_2')$ and $D_s=\{(h_1,h_2)\in \mathbb{R}^2|h_1<s \mbox{ and } h_2>s-h_1\}$. Also in what follows, for a random variable $Y$
we denote $f_{Y}$ its probability distribution. 
Using these notations and the fact that $\xi$ is independent of $X_t$, we can write: 
\begin{align*}
\Pr(X_{t+s}=y,\xi\in D_s |X_t=x)&=\int_{D_s}f_{(X_{t+s},\xi)| X_t=x}(y,h) \mathrm{d} h\\
&=\int_{D_s}f_{X_{t+s}| \xi=h, X_t=x}(y)f_{\xi}(h) \mathrm{d} h\\
&=\int_0^s\int_{s-h_1}^{+\infty} q_{\beta_{t+h_1}}(x,y) e^{-h_1} e^{-h_2}\mathrm{d}h_1 \mathrm{d}h_2
\end{align*}

The previous equality yields:

\begin{equation}\label{term2}
\Pr(X_{t+s}=y,~H_{t+s}-H_{t}= 1|X_{t}=x)=e^{-s}\int_{0}^{s}  q_{\beta_{t+h_1}}(x,y) \mathrm{d}h_1.
\end{equation}

In the following we use the classical $\mathcal{O}(.)$ and $o(.)$ notations: for all functions $f$ and $g$ defined on some subset of $\mathbb{R}$,  
\begin{itemize}
	\item  $f(x)=\mathcal{O}(g(x))\mbox{ as }x\to 0^+ \Longleftrightarrow \exists \sigma,x_0>0,  |f(x)| \le \; \sigma |g(x)|\mbox{ for all }0<x \leq x_0$
	\item $f(x)=o(g(x))\mbox{ as }x\to 0^+ \Longleftrightarrow \underset{x\to 0^+}{\lim}\frac{f(x)}{g(x)}=0.$ 
\end{itemize}
For the third term we can see that:
$$\Pr(X_{t+s}=y,~H_{t+s}-H_{t}\ge 2|X_{t}=x) \le \Pr(H_{t+s}-H_{t}\ge 2)\le  \Pr( H_{s}\ge 2)$$ 
Since $H_s$ is a Poisson Process of parameter $1$, one can check that for all $s$ close to zero we have that $\Pr( H_{s}\ge 2)=1-\Pr( H_{s}= 0)-\Pr( H_{s}= 1)=1-e^{-s}-se^{-s}= \mathcal{O}(s^2)$.

This implies  that when $s$ is close to zero, the probability that the process goes from $x$ to $y$ between $t$ and $t+s$, with more than one jump  is small in comparison to $s$:
\begin{equation}\label{term3}
\Pr(X_{t+s}=y,~H_{t+s}-H_{t}\ge 2|X_{t}=x) =\mathcal{O}(s^2)
\end{equation}
Putting all the terms together and replacing them in \eqref{defgenerator}, we can rewrite the infinitesimal generator as follows:

\begin{align*}
L_{\B_t}f(x)&=\lim\limits_{s\to 0} \frac{1}{s}\quad \biggl[\sum\limits_{y\in E}f(y)\biggl[\delta_{x}(y)e^{-s}+e^{-s}\int_{0}^{s}  q_{\beta(t+\tau)}(x,y) d\tau\biggr]-f(x)\biggr]\\
&+\lim\limits_{s\to 0} \frac{1}{s}\sum\limits_{y\in E}f(y)\Pr(X_{t+s}=y, H_{t+s}-H_{t}\ge 2|X_{t}=x)\\
\end{align*}
Using the fact that $f$ is bounded, $E$ finite and the upper bound given by \eqref{term3}, one can easily check that the second term is zero. Hence we obtain:
\begin{align*}
L_{\B_t}f(x)&=\lim\limits_{s\to 0} \frac{1}{s}\quad\left[e^{-s}\left[f(x)+\sum\limits_{y\in E}f(y)\int_{0}^{s} q_{\beta(t+\tau)}(x,y) d\tau\right] -f(x)\right]\\
&=\lim_{s\to 0} \frac{f(x)(e^{-s}-1)}{s} +\lim_{s\to 0} \frac{e^{-s} }{s}\left(\sum_{y\in E}f(y)\int_{0}^{s} q_{\beta(t+\tau)}(x,y) d\tau\right)
\end{align*}

Noting the fact that $q_{\beta_t}$ is continuous with respect to $t$ and the following identity 
$$ e^{-s}=1-s+\mathcal{O}(s^2),$$
we easily obtain the simplest form for the infinitesimal generator of the inhomogeneous Markov chain:
\begin{equation}\label{classic generator}
L_{\B_t}f(x)=\sum\limits_{y\in E}\biggl(f(y)-f(x)\biggr)q_{\beta_t}(x,y).
\end{equation}
We can remark that the explicit form of the transition probability $q_{\beta_t}$ does not appear in the proof, hence the result is completely general. The only necessary property of this transition probability is its continuity with respect to $t$. 

The fact that $n_t$ and $\beta_t$ are continuous functions ensures the continuity of transition probability  $\tilde{q}_{\beta_t}$, defined in \eqref{eq:def_qtilde} . Therefore, following the same argument,one can deduce (\ref{nsa generator}).  Here we can see the relevance of the randomness of $N_t$. An increasing  deterministic sequence would generate a discontinuous  $\tilde{q}_{\beta_t}$ and would make difficult the use of derivations above.

\section{Proof, part 2: Generators comparison}\label{sec:gen_diff}
The fact that for a temperature schedule that decreases slowly enough, the process generated by the classical Simulated Annealing converges to the set of global minima of $J$ is well known. The Noisy Simulated Annealing is a similar algorithm, built on the same principles except that the values of the function $J$ are replaced by an estimation each time its computation is needed. Therefore a tight relation exists between both approaches. Furthermore, as we will show in this section, for a well chosen couple $(\beta_t,n_t)$ the generators of the two algorithms will be 'close' at large times. This is a key element of the proof as it will imply a first condition for the ratio $\beta_t/n_t$.     

Using the relations given by \eqref{classic generator} and \eqref{nsa generator},  the quantity of interest is:
\begin{align*}
\tilde{L}_{\beta_t}f(x)=L_{\beta_t}f(x) + \sum_{y\in E}(f(y)-f(x))(\tilde{q_{\beta_t}}-q_{\beta_t})(x,y).
\end{align*}
Hence quantifying the difference between the two generators can be reduced to bounding the difference between the two probability transitions $q_{\beta_t}$ and $\tilde{q}_{\beta_t}$.  
Thus the main result of this section is the following lemma.
\begin{lem}\label{bounds}Let $\beta_t/\sqrt{n_t}\underset{\infty}{\to} 0$. There exist two functions $\ei_t$ and $\es_t$ such that 
	$$\pourtout{t}{\Rp}\pourtout{x}{E'}\pourtout{y}{E} \quad \ei_t q_{\Bt}(x,y)\le(\tilde{q}_{\beta_t}-q_{\beta_t})(x,y)\le \es_t q_{\Bt}(x,y)$$
	and $$ \lim\limits_{t\to+\infty} \ei_t=\lim\limits_{t\to+\infty}\es_t=0.$$
\end{lem}	
Before going into the proof of this lemma, we present some preliminaries. First it can be noticed that for all $x,y\in E$, $x\ne y$ we have:

\begin{align*}
&(\tilde{q_{\beta_t}}-q_{\beta_t})(x,y) \\
&=q_{\beta_t}(x,y)\left(\frac{\tilde{q_{\beta_t}}}{q_{\beta_t}}-1\right)(x,y)\\
& =q_{\beta_t}(x,y) \left(\mathbb{E}_{N_t}\mathbb{E}_{\omega_1,...,\omega_{N_t}}\left(\frac{e^{-\frac{\beta}{N_t}\lfloor \sum_{i=1}^{N_t} U(y,\omega_i)-U(x,\omega_i)\rfloor_{+}}}{e^{-\beta_t\lfloor \mathbb{E}(U(y,\Omega))-\mathbb{E}(U(x,\Omega)) \rfloor_{+}}}-1\right)\right).
\end{align*}

Unless specified otherwise, in this section we always consider $x\ne y$ . The case $x=y$ is handled at the end of the section. To simplify the notations we denote $X_i^{x,y}:= U(y,\omega_i)-U(x,\omega_i)-\mathbb{E}(U(y,\omega_i)-U(x,\omega_i))$ and $K^{x,y}=\mathbb{E}(U(y,\omega_i)-U(x,\omega_i))$. Hence,

\begin{align*}
&(\tilde{q_{\beta_t}}-q_{\beta_t})(x,y) \\
& =q_{\beta_t}(x,y) \left(\mathbb{E}_{N_t}\mathbb{E}_{\omega_1,...,\omega_{N_t}}\left(e^{- \beta_t\left(\lfloor \frac{1}{N_t}\sum_{i=1}^{N_t} X_i^{x,y} +  K^{x,y}\rfloor_{+}+\lfloor K^{x,y}\rfloor_{+} \right)}-1\right)\right).
\end{align*}
Noticing that, 
\begin{align*}
\forall~a,b\in \mathbb{R},\quad -|a|\le -\lfloor a+ b \rfloor_{+} + \lfloor b \rfloor_{+}\le |a|,
\end{align*}
we obtain the following bounds for $(\tilde{q_{\beta_t}}-q_{\beta_t})(x,y)$:

\begin{align}\label{qbounds}
& \mathbb{E}_{N_t}\mathbb{E}_{\omega_1,...,\omega_{N_t}}\left(e^{|\frac{\beta_t}{N_t}\sum_{i=1}^{N_t} X_i^{x,y}|}-1\right)\ge (\frac{\tilde{q_{\beta_t}}-q_{\beta_t}}{q_{\beta_t}})(x,y) \ge  \mathbb{E}_{N_t}\mathbb{E}_{\omega_1,...,\omega_{N_t}}\left(e^{-|\frac{\beta_t}{N_t}\sum_{i=1}^{N_t} X_i^{x,y}|}-1\right).
\end{align}
In order to obtain a bound for the expectation of a function of $N_t$ we need an estimation of the  probability that $N_t$ takes values 'far' from its expectation.  	

\medskip
\begin{lem}\label{poisson}\hclement{d\'{e}finir $a$ avant}
	There exist $\delta\in(0,1)$ and $a=\lvert (1-\delta)\left(1-\log(1-\delta)\right)-1\rvert$ such that for all $t>0$ we have:
	$$\proba(N_t\le (1-\delta)n_t)\le e^{-a n_t}.$$
\end{lem}
{\bf Proof.}
We remind the reader that, as mentioned in the definition \asumref{asum:poisson}, at a fixed time $t$  the process  $N_t$ can be written as $1+H$, where $H$ is a Poisson random variable of parameter $n_t$.
\medskip

Fix $t\in\mathbb{R}$ and $\delta\in (0,1)$. We provide an upper bound for $\mathbb{P}(N_t\le (1-\delta)n_t)$ using the Cramer-Chernoff method.

\medskip

First we see that 
for any $\lambda>0$, applying Markov's inequality we have:
\begin{align*}
\mathbb{P}(N_t\le (1-\delta)n_t)=\proba(e^{-\lambda N_t}>e^{\lambda (\delta-1)n_t})\le\dfrac{\expen[e^{-\lambda N_t}]}{e^{\lambda (\delta-1)n_t}}..
\end{align*}

For $t$ fixed $N_t-1$ has the distribution of a Poisson random variable of parameter $n_t$.  Therefore by direct computations we have:
\begin{align*}
\expen[e^{-\lambda N_t}]&=\sum_{k>0}(e^{-\lambda(k+1)} e^{-n_t}\dfrac{n_t^k}{k!})\\
&=e^{-n_t-\lambda}\sum_{k>0}\dfrac{(e^{-\lambda}n_t)^k}{k!}\\
&=e^{-n_t-\lambda} e^{e^{-\lambda}n_t}
\end{align*}

Putting all these elements together yields: 
$$\proba(N_t\le (1-\delta)n_t)\le \exp([-\lambda(\delta-1)-1+e^{-\lambda}]n_t)$$
The idea is to choose $\lambda$ and $\delta$ in order to obtain the smallest possible value for $-\lambda(\delta-1)-1+e^{-\lambda}$. For  $\delta\in(0,1)$, the minimum is reached at $\lambda=-\log(1-\delta)$ which is strictly positive.  For such $\lambda$ and $\delta$, we denote $a=\lvert (1-\delta)\left(1-\log(1-\delta)\right)-1\rvert$ and conclude the proof. 

\hfill\BlackBox\\

Now we have what we need in order to start the proof of Lemma \ref{bounds}.

{\bf Proof of Lemma \ref{bounds}.}

First, recall that if $y\in E\backslash E'$ then $\forall x \in E',\quad \tilde{q}_{\beta_t}(x,y)=q_{\beta_t}(x,y)=0$. Hence Lemma \eqref{bounds} is trivially verified for $x \in E$ and $y\in E\backslash E'$.\\
Considering the inequalities given by \eqref{qbounds}, the proof can  be divided in two parts by studying separately the upper bound $\mathbb{E}_{N_t}\mathbb{E}_{\omega_1,...,\omega_{N_t}}\left(e^{|\frac{\beta_t}{N_t}\sum_{i=1}^{N_t} X_i^{x,y}|}\right)$ and the lower bound $\mathbb{E}_{N_t}\mathbb{E}_{\omega_1,...,\omega_{N_t}}\left(e^{-|\frac{\beta_t}{N_t}\sum_{i=1}^{N_t} X_i^{x,y}|}\right)$.\\	

\paragraph{Upper bound}

We have:
$$(\tilde{q_{\beta_t}}-q_{\beta_t})(x,y)\le q_{\beta_t}(x,y) \left(\mathbb{E}_{N_t}\mathbb{E}_{\omega_1,...,\omega_{N_t}}\left(e^{|\frac{\beta_t}{N_t}\sum_{i=1}^{N_t} X_i^{x,y}|}-1\right)\right).
\quad$$

First we will provide an estimate of $\mathbb{E}_{\omega_1,...,\omega_{N_t}}\left[e^{|\frac{\beta_t}{N_t}\sum_{i=1}^{N_t} X_i^{x,y}|}|N_t\right]$ for all $N_t$.

We start by rewriting this expectation as: 
\begin{align*}
\mathbb{E}_{\omega_1,...,\omega_{N_t}}& \left[e^{|\frac{\beta_t}{N_t}\sum_{i=1}^{N_t} X_i^{x,y}|}|N_t\right]=\int_{\mathbb{R}_+}\Pr\left(e^{|\frac{\beta_t}{N_t}\sum_{i=1}^{N_t} X_i^{x,y}|}>u|N_t \right)du\\
&= \int_{\mathbb{R}_+}\Pr\left(\left|\sum_{i=1}^{N_t} X_i^{x,y}\right|>\frac{\log(u)N_t}{\beta_t}|N_t \right)du.\\
\end{align*}
Since $X_i^{x,y}= U(y,\omega_i)-U(x,\omega_i)-\mathbb{E}_{N_t}\mathbb{E}_{\omega_i}(U(y,\omega_i)-U(x,\omega_i))$ is a centered random variable and $U$ is bounded on $E'$ (see the definition of \asumref{asum:Ubounded}) there exists $\sigma$ such that $|X_i^{x,y}|\le \sigma$ (for example set $\sigma=2\maxofU$), almost surely for all $i$. Therefore $(X_i^{x,y})_{1\le i\le N_t}$ are sub-gaussian random variables (\cite{boucheron2013concentration}) with variance factor $\sigma^2$, \textit{i.e.}, $$\forall u\ge 0,~ \max\left( \Pr(X_i^{x,y}>u),\Pr(-X_i^{x,y}>u) \right)\le e^{-\frac{u^2}{2\sigma^2}}.$$ 
Considering that $(X_i^{x,y})_{i\le N_t}$ is a sequence of independent sub-Gaussian variables, their sum is still a sub-Gaussian variable.  As $\Var(X_i^{x,y})\le \sigma^2$ for all $i$ we have that $\Var\left(\sum_{i=1}^{N_t} X_i^{x,y} \right)\le N_t \sigma^2$ and therefore:
$$\forall u\ge 0,~ \max\left(\Pr\left(\sum_{i=1}^{N_t} X_i^{x,y} > u\right),\Pr\left( -\sum_{i=1}^{N_t} X_i^{x,y} >u\right)\right)\le e^{-\frac{u^2}{2\sigma^2N_t}}.$$ 

For more details about sub-gaussian variables we refer to \cite{boucheron2013concentration}. We use this property of concentration in order to get an estimate of the expectation:

\begin{align*}
\mathbb{E}_{\omega_1,...,\omega_{N_t}}& \left[e^{|\frac{\beta_t}{N_t}\sum_{i=1}^{N_t} X_i^{x,y}|}|N_t\right]\\ &= \int_{0}^{1}\Pr\left(\left|\sum_{i=1}^{N_t} X_i^{x,y}\right|>\frac{\log(u)N_t}{\beta_t}|N_t \right)du+ \int_{1}^{+\infty}\Pr\left(\left|\sum_{i=1}^{N_t} X_i^{x,y}\right|>\frac{\log(u)N_t}{\beta_t}|N_t \right)du\\
&\le 1+ 2\int_{1}^{+\infty}e^{-\frac{1}{2\sigma^2}\left(\frac{\log(u)\sqrt{N_t}}{\beta_t}\right)^2}du.
\end{align*}
Using a simple variable substitution $s=\frac{\log(u)}{\lambda_t}$ with $\lambda_t=\frac{\sigma\beta_t}{\sqrt{N_t}}$, we get:
\begin{align*}
\mathbb{E}_{\omega_1,...,\omega_{N_t}} \left[e^{|\frac{\beta_t}{N_t}\sum_{i=1}^{N_t} X_i^{x,y}|}|N_t\right]&\le 1+ 2\lambda_t\int_{0}^{+\infty}e^{\lambda_ts}e^{-\frac{s^2}{2}}du\\
&\le 1+ 2\lambda_te^{\frac{\lambda_t^2}{2}}\int_{0}^{+\infty}e^{-\frac{(s-\lambda_t)^2}{2}}ds\\
&\le 1+ 2\lambda_te^{\frac{\lambda_t^2}{2}}\int_{-\lambda_t}^{+\infty}e^{-\frac{u^2}{2}}du\\
&\le 1+ 2\sqrt{2\pi}\lambda_te^{\frac{\lambda_t^2}{2}}\Pr(G>-\lambda_t)\\
&\le 1+ 2\sqrt{2\pi}\lambda_te^{\frac{\lambda_t^2}{2}}(1-\Pr(G>\lambda_t)).
\end{align*}
where $G$ is a standard Gaussian. Thanks to the Taylor formula, we know that there exists a constant $0<\theta<1$, such that:
\begin{align*}
\Pr(G>\lambda_t)&=\Pr(G>0)+\lambda_t\frac{e^{\frac{-(\theta\lambda_t)^2}{2}}}{\sqrt{2\pi}}\\
&= \frac{1}{2}+\lambda_t\frac{e^{\frac{-(\theta\lambda_t)^2}{2}}}{\sqrt{2\pi}}\\
&\ge \frac{1}{2}+\lambda_t\frac{e^{\frac{-(\lambda_t)^2}{2}}}{\sqrt{2\pi}}.
\end{align*}
This leads to: 
\begin{align*}
\mathbb{E}_{\omega_1,...,\omega_{N_t}} \left[e^{|\frac{\beta_t}{N_t}\sum_{i=1}^{N_t} X_i^{x,y}|}|N_t\right]&\le 1+ (\sqrt{2\pi}e^{\frac{\lambda_t^2}{2}})\lambda_t.
\end{align*}
Thus, replacing $\lambda_t$ by its definition we see that we  need an estimate of: $$\expen\left[1+ (\sqrt{2\pi}e^{\frac{\sigma^2\beta_t^2}{2N_t}})\frac{\sigma\beta_t}{\sqrt{N_t}}\right].$$ 
In order to simplify the notations we denote: $g_t= \sigma \beta_t$. Using the bound given by Lemma $\ref{poisson}$ we get:
\begin{align*}
\expen&\left[1+ \sqrt{2\pi}e^{\frac{g_t^2}{2\sqrt{N_t}}}\frac{g_t}{\sqrt{N_t}}\right] \\
&=1+\expen\left[ \sqrt{2\pi}e^{\frac{g_t^2}{2\sqrt{N_t}}}\frac{g_t}{\sqrt{N_t}}\right]\\
&\le 1+\expen\left[ \sqrt{2\pi}e^{\frac{g_t^2}{2\sqrt{N_t}}}\frac{g_t}{\sqrt{N_t}} (\mathds{1}_{[1,(1-\delta)n_t]}+\mathds{1}_{[(1-\delta)n_t,+\infty)})\right]\\
&\le 1+\sqrt{2\pi}g_t\left[\frac{e^{\frac{g_t^2}{2(1-\delta)n_t}}}{\sqrt{(1-\delta)n_t}}+e^{g_t^2-an_t}\right]
\end{align*}
\medskip
Finally we have obtained that under assumptions of Section \ref{assumption_subsec}:
\begin{align*}
(\tilde{q_{\beta_t}}-q_{\beta_t})(x,y)&\le q_{\beta_t}(x,y) \left(\mathbb{E}_{N_t}\mathbb{E}_{\omega_1,...,\omega_{N_t}}\left(e^{|\frac{\beta_t}{N_t}\sum_{i=1}^{N_t} X_i^{x,y}|}-1\right)\right)\\
&\le q_{\beta_t}(x,y) \sqrt{2\pi}g_t\left[\frac{e^{\frac{g_t^2}{2(1-\delta)n_t}}}{\sqrt{(1-\delta)n_t}}+e^{g_t^2-an_t}\right]
\end{align*}
Hence it is natural to define $\es_t$ as: $$\es_t=\sqrt{2\pi}\beta_t\sigma\left[\frac{e^{\frac{\beta_t^2\sigma^2}{2(1-\delta)n_t}}}{\sqrt{(1-\delta)n_t}}+e^{\beta_t^2\sigma^2-an_t}\right].$$	
Since  $\beta_t/\sqrt{n_t}\underset{\infty}{\to} 0$ one can check that $\es_t$ goes to $0$ when $t$ goes to infinity. Here we can see once more the importance of the balance between the two parameters $\beta_t$ and $n_t$.   

		\paragraph{Lower bound}
		Considering the left-hand side of \eqref{qbounds} we have: $$ (\tilde{q_{\beta_t}}-q_{\beta_t})(x,y) \ge 
		q_{\beta_t}(x,y) \left(\mathbb{E}_{N_t}\mathbb{E}_{\omega_1,...,\omega_{N_t}}\left(e^{-|\frac{\beta_t}{N_t}\sum_{i=1}^{N_t} X_i^{x,y}|}-1\right)\right).$$
		In a similar way we start by obtaining a lower bound for $\mathbb{E}_{\omega_1,...,\omega_{N_t}}\left(e^{-|\frac{\beta_t}{N_t}\sum_{i=1}^{N_t} X_i^{x,y}|}|N_t\right)$ and after we improve it  using the probabilistic properties of $N_t$.
		\medskip		
				First observe that:
		\begin{align*}
		\mathbb{E}_{\omega_1,...,\omega_{N_t}}& \left[e^{-|\frac{\beta_t}{N_t}\sum_{i=1}^{N_t} X_i^{x,y}|}|N_t\right]=\int_{\mathbb{R}_+}\Pr\left(e^{-|\frac{\beta_t}{N_t}\sum_{i=1}^{N_t} X_i^{x,y}|}>u|N_t \right)du\\
		&= \int_{\mathbb{R}_+}\Pr\left(-\left|\sum_{i=1}^{N_t} X_i^{x,y}\right|>\frac{\log(u)N_t}{\beta_t}|N_t \right)du\\
		&= \int_{0}^{1}\Pr\left(\left|\sum_{i=1}^{N_t} X_i^{x,y}\right|<\frac{-\log(u)N_t}{\beta_t}|N_t \right)du\\
		&= \int_{0}^{1}1-\Pr\left(\left|\sum_{i=1}^{N_t} X_i^{x,y}\right|>\frac{-\log(u)N_t}{\beta_t}|N_t \right)du\\
		&\ge 1- 2\int_{0}^{1}e^{-\frac{N_t}{2\sigma^2}\left(\frac{-\log(u)}{\beta_t}\right)^2}du
		\end{align*}
		
		Again, this is due to the fact that the sum of  $X_i^{x,y}$ is sub-Gaussian with variance factor $N_t\sigma^2$.
		
		Using the same variable substitution as above: $s=\frac{\log(u)}{\lambda_t}$ with $\lambda_t=\frac{\sigma\beta_t}{\sqrt{N_t}}$, we get:
		\begin{align*}
		\mathbb{E}_{\omega_1,...,\omega_{N_t}} \left[e^{-|\frac{\beta_t}{N_t}\sum_{i=1}^{N_t} X_i^{x,y}|}|N_t\right]&\ge 1- 2\lambda_t\int_{-\infty}^{0}e^{\lambda_ts}e^{-\frac{s^2}{2}}du\\
		&\ge 1- 2\lambda_te^{\frac{\lambda_t^2}{2}}\int_{-\infty}^{0}e^{-\frac{(s-\lambda_t)^2}{2}}ds\\
		&\ge 1- 2\lambda_te^{\frac{\lambda_t^2}{2}}\int_{-\infty}^{-\lambda_t}e^{-\frac{u^2}{2}}du\\
		&\ge 1- 2\sqrt{2\pi}\lambda_te^{\frac{\lambda_t^2}{2}}\Pr(G>\lambda_t)
		\end{align*}
		where $G$ is a standard $N(0,1)$ Gaussian. As seen before there exists some $0<\theta<1$, such that:
		\begin{align*}
		\Pr(G>\lambda_t)&\ge \frac{1}{2}+\lambda_t\frac{e^{\frac{-(\lambda_t)^2}{2}}}{\sqrt{2\pi}}.
		\end{align*}
		This leads to 
		\begin{align*}
		\mathbb{E}_{\omega_1,...,\omega_{N_t}} \left[e^{|\frac{\beta_t}{N_t}\sum_{i=1}^{N_t} X_i^{x,y}|}|N_t\right]&\ge 1- (\sqrt{2\pi}e^{\frac{\lambda_t^2}{2}})\lambda_t.
		\end{align*}
		This expression has exactly the symmetric form to the one obtained in the upper bound part. Thus, the lower bound is obtained the same way as the upper bound. We directly get:
		
		\begin{align*}
		\expen&\left[1- (\sqrt{2\pi}e^{\frac{g_t^2}{2N_t}})\frac{g_t}{\sqrt{N_t}}\right] \\
		&\ge 1-\sqrt{2\pi}g_t\left[\frac{e^{\frac{g_t^2}{2(1-\delta)n_t}}}{\sqrt{(1-\delta)n_t}}+e^{g_t^2-an_t}\right]
		\end{align*}

		Now we define $\ei_t$: $$\ei_t=-\sqrt{2\pi}\beta_t\sigma\left[\frac{e^{\frac{\beta_t^2\sigma^2}{2(1-\delta)n_t}}}{\sqrt{(1-\delta)n_t}}+e^{\beta_t^2\sigma^2-an_t}\right].$$ It is easy to see that $\ei_t$ goes to $0$ when $t$ goes to infinity as soon as $\beta_t/\sqrt{n_t}\underset{\infty}{\to} 0$. This completes the proof of Lemma \ref{bounds}.
	\hfill\BlackBox\\
	
	\section{Proof, last part: rate of convergence in the general case}\label{sec:gronwall}
	We first complete the proof of convergence as stated in \thmref{thm:maintheorem} and then deduce the convergence rate  (\thmref{thm:conv_speed}) from it. This enables us to provide an upper bound on the minimal number of cost function evaluations in \seqref{comp_complex}.
	\subsection{Proof of \thmref{thm:maintheorem} }
	The proof of \thmref{thm:maintheorem} follows the roadmap of Holley and Strook \cite{Holley88Simu} and relies on the use of the Gr\"onwall lemma. We derive a differential inequality for the $L^2_{\mu_{\beta_t}}$-norm of the density measure of the NSA process with respect to $\mub$ and deduce an integrated version of it using the lemma. We then show that bounding the $L^2_{\mu_{\beta_t}}$-norm of this density implies the convergence of the process to the optimal state space $\Xe$.\\

	\noindent 
	{\bf Proof. [Proof of \thmref{thm:maintheorem}]}~
		Our goal is to show that when $t$ goes to infinity, the noisy simulated annealing gets ``close enough'' to the classical simulated annealing. Therefore we denote by $f_t$ the Radon-Nikodym derivative of the probability density of the noisy simulated annealing process $\tilde{X}_t$ with respect to the Gibbs measure $\mu_{\Bt}$, \textit{i.e.}:
		\begin{equation}\label{density_def}
f_t=\frac{\mathrm{d} m_t }{\mathrm{d} \mu_{\Bt}}
		\end{equation}  where $m_t$ is the distribution of $(\tilde{X}_s)_{s\ge 0}$ at time $t$. 
		A first remark is that $f_t(x)=0$ for all $t\ge 0$ and all $x\in E\setminus E'$, since our process, by construction does not accept states out of $E'$.\\
		Using the results obtained in \seqref{sec:inf_gen} one can see that $\mathbb{R}_{+}\ni t\to\tilde{L}_{\beta_t}$ is continuous and therefore the semi-group $(P_{s,t})_{0\le s \le t}$ is smooth. 
		Also by their definition the operators $(P_{s,t})_{0\le s \le t}$ are linear and have the following semi-group property:  $P_{s,t+h}=P_{s,t} \circ P_{t,t+h} $, for all $0\le s <t $ and $h>0$. 
		Hence, for all $0\le s\le t$, we have:		
		\begin{equation}\label{Kolmogorov}
		\frac{\mathrm{d}}{\mathrm{d}t} P_{s,t}= P_{s,t}L_t.
		\end{equation}
		For details about the infinitesimal generator see Section 1.4. of 	 \cite{bakry2013analysis}.\\

		As shown in \eqref{ftbound}, bounding the $L^2$-norm of $f_t$ w.r.t. $\mub$, \textit{i.e.}, $\|f_t\|_{\mub}$,  ensures convergence of the NSA algorithm. However it does not provide enough information about the convergence of $m_t$ to $\mub$ to deduce a fine convergence rate. This is why we study the evolution of $\|f_t-1\|_{\mub}$ which controls the distance between the two measures. If this quantity is bounded then we obtain the convergence of the NSA algorithm. If moreover it converges to zero, it implies a stronger convergence rate.
		 In order to prove that, we deduce a differential inequality for $\|f_t-1\|^2_{\mub}$. We start by computing its derivative:
		\begin{align*}
		\dt \|f_t-1\|^2_{\mub}=\dt \|f_t\|^2_{\mub}=& \dt \sum_{x\in E} f_t^2(x) \mub(x)\\
		&=2\sum_{x\in E} f_t(x) \dt \left[\frac{m_t}{\mub}\right](x) \mub(x)+\sum_{x\in E} f_t(x)\dt \mub(x).
		\end{align*}
		Using the backward Kolmogorov equation given by \eqref{Kolmogorov}, for the first term we have:  
		\begin{align}
		\sum_{x\in E} f_t(x) \dt \left[\frac{m_t}{\mub}\right](x) \mub(x)
		&=\sum_{x\in E} f_t(x) \dt \left[\frac{m_t}{\mub}\right](x) \mub(x)+\sum_{x\in E} f_t(x)\dt \mub(x) \label{lemma6Holley}\\
		&=\sum_{x\in E} \left[\tilde{L}_{\beta_t} f_t(x) \right] m_t(x) -\sum_{x\in E} f_t(x) \frac{m_t}{\mub}(x)\dt \mub(x). \nonumber
		\end{align}
		Denote $\langle J\rangle_{\mub}:=\int J d\mu_{\beta_t}$  the mean of $J$ with respect to $\mu_{\beta_t}$. One can check that:
		$$\dt \mub (x)= -\beta_t'\left[J(x)-\langle J\rangle_{\mub}\right] \mub(x). $$
		Thus, we easily obtain the following equality:	
		\begin{equation}\label{densderiv}
		\dt\|f_t-1\|_{ \mu_{\beta_t}}^2=2\sum_{x\in E}f_t(x)(\tilde{L}_{\beta_t}f_t)(x) \mu_{\beta_t}(x)+ \beta_t'\sum_{x\in E}(J(x)-\langle J\rangle_{\mub})f_t^2(x)\mu_{\beta_t}.
		\end{equation}
		First, we focus on the first term of the right hand side of \eqref{densderiv}. Since we try to control the generator of the noisy simulated annealing by the generator of the classical one, it is natural to write $\tilde{L}_{\beta_t}$ as $L_{\beta_t}+\tilde{L}_{\beta_t}-L_{\beta_t}$. This comparison leads to the computation:
		
		\begin{align*}
		&\sum_{x\in E}f_t(x)(\tilde{L}_{\beta_t}f_t)(x) \mu_{\beta_t}(x)\\
		&	=\sum_{x}f_t(x)(L_{\beta_t}f_t)(x) \mu_{\beta_t}(x)+ \sum_{x\in E}f_t(x)\left(\sum_{y\in E}(f_t(y)-f_t(x))(\tilde{q_{\beta_t}}-q_{\beta_t})(x,y)\right)\mu_{\beta_t}(x).	
		\end{align*}
		We rewrite the last part of the second term using Lemma \ref{bounds}:
		\begin{align*}
		&\sum_{y\in E}(f_t(y)-f_t(x))(\tilde{q_{\beta_t}}-q_{\beta_t})(x,y)\\
		&=\sum_{y\in E}f_t(y)(\tilde{q_{\beta_t}}-q_{\beta_t})(x,y)-\sum_{y\in E}f_t(x))(\tilde{q_{\beta_t}}-q_{\beta_t})(x,y)\\
		&\le \es_t\sum_{y\in E}f_t(y)q_{\beta_t}(x,y)-\es_t\sum_{y\in E}f_t(x)q_{\beta_t}(x,y)+\es_t\sum_{y\in E}f_t(x)q_{\beta_t}(x,y)-\ei_t\sum_{y\in E}f_t(x)q_{\beta_t}(x,y)\\
		&	\le \es_t\sum_{y\in E}(f_t(y)-f_t(x))q_{\beta_t}(x,y)+ (\es_t-\ei_t)f_t(x)\\
		&	\le \es_t L_{\beta_t}f_t(x)+ (\es_t-\ei_t)f_t(x).		
		\end{align*}
		Inserting this in the previous inequality, we get: 
		\begin{align*}
		&	\sum_{x\in E}f_t(x)(\tilde{L}_{\beta_t}f_t)(x) \mu_{\beta_t}(x)\\
		&	\le(1+\es_t)\sum_{x}f_t(x)(L_{\beta_t}f_t)(x) \mu_{\beta_t}(x)+(\es_t-\ei_t) \sum_{x\in E}f_t^2(x)\mu_{\beta_t}(x)\\
		&		\le(1+\es_t)\sum_{x}f_t(x)(L_{\beta_t}f_t)(x) \mu_{\beta_t}(x)+(\es_t-\ei_t) \sum_{x\in E}(f_t^2(x)-1)\mu_{\beta_t}(x)+(\es_t-\ei_t).
		\end{align*}
		Therefore, using \eqref{densderiv}, we obtain the following inequality:				
		\begin{align*}
		&\frac{d}{dt}\|f_t-1\|_{ \mu_{\beta_t}}^2 \\
		&\le 2 \left[(1+\es_t)\sum_{x}f_t(x)(L_{\beta_t}f_t)(x) \mu_{\beta_t}(x)+(\es_t-\ei_t) \sum_{x\in E}(f_t^2(x)-1)\mu_{\beta_t}(x)+(\es_t-\ei_t)\right]\\
		&\quad\quad +\beta_t'\sum_{x\in E}(J(x)-\langle J\rangle_{\mub})f_t^2(x)\mu_{\beta_t}\\
		&\le 2 \left[(1+\es_t)\sum_{x}f_t(x)(L_{\beta_t}f_t)(x) \mu_{\beta_t}(x)+(\es_t-\ei_t) \sum_{x\in E}(f_t^2(x)-1)\mu_{\beta_t}(x)+(\es_t-\ei_t)\right]\\
		&\quad\quad +\beta_t'\sum_{x\in E}(J(x)-\langle J\rangle_{\mub})(f_t(x)-1)^2\mu_{\beta_t}(x)+2\beta_t'\sum_{x\in E}(J(x)-\langle J\rangle_{\mub})(f_t(x)-1)\mu_{\beta_t(x)}.
		\end{align*}
		In order to deal with the first sum we use an estimate of the spectral gap of $L_{\beta_t} $. This is provided by Theorem 2.1 of Holley and Strook\cite{Holley88Simu}.\\
		
		\noindent
		\begin{theorem}[Holley and Strook 88]\label{thHolley}
			Under assumptions of \ref{assumption_subsec}, there exist two positive constants $0<c\le C<+\infty$ such that $\pourtout{\beta}{\Rp}$
			$$ce^{-\beta m^{\star}}\le \gamma(\beta) \le Ce^{-\beta m^{\star}}$$
			where $\gamma(\beta)= \inf\{-\int \phi L_{\beta}\phi~d\mu_{\beta} : \| \phi \|_{\mu_{\beta}}=1~\mbox{and}~\int \phi d\mu_{\beta}=0 \}$ and $m^{\star}$ is the maximum depth of a well containing a local minimum defined in $\eqref{definition of m}$.
		\hfill\BlackBox
		\end{theorem}
		{\bf Remark}{
			The constant $m^{\star}$ is always strictly positive as soon as the function has a strict local minimum  that is not global. This is generally the case in our setting. Also, we always have $m^{\star}\le \maxofU$.
		}
		
		{\bf  Following of the Proof.} Using the definition of $f_t$, one can see that $\int f_t d\mu_{\beta_t}=1$, hence applying the theorem for $\phi=\dfrac{f_t-1}{\|f_t-1\|_{ \mu_{\beta_t}}}$ gives:
		$$-\sum_{x}\phi(L_{\beta_t}\phi)(x) \mu_{\beta_t}(x)\geq ce^{-\beta_t m^{\star}}. $$
		This and the definition of $L_{\beta_t}$ imply:
		$$\sum_{x}f_t(x)(L_{\beta_t}f_t)(x) \mu_{\beta_t}(x)\le -ce^{-\beta_t m^{\star}} \|f_t-1\|_{ \mu_{\beta_t}}^2.$$
		$J$ is a positive function bounded by $\maxofU$ on $E'$. For all $x \in E\setminus E'$, the only points where $J>\maxofU$, we have that $J(x)=+\infty$ and therefore $\mu_{\beta_t}(x)=0$. This implies that for all measurable functions $g$, $$\displaystyle\sum_{x\in E}(J(x)-\langle J\rangle_{\mub})g(x)\mu_{\beta_t}\le \maxofU \|g\|_{\mu_{\beta_t}}.$$
		Putting all these terms together gives:
		\begin{align}
		\frac{d}{dt}\|f_t-1\|_{ \mu_{\beta_t}}^2\le& ~2 \left[-c e^{-\beta_tm^{\star}}(1+\es_t)+(\es_t-\ei_t)+\frac{\maxofU}{2}\beta_t'\right]\|f_t-1\|_{ \mu_{\beta_t}}^2\\
		&+2\maxofU\beta_t'\|f_t-1\|_{ \mu_{\beta_t}}\nonumber\\
		&+2(\es_t-\ei_t).\nonumber
		\end{align}
		We denote $u_t=\|f_t-1\|_{ \mu_{\beta_t}}^2$. Considering the fact that $\es_t$ is a positive function we have: 
		\begin{align}\label{gronineq}
		u_t'\le& ~2 \left[-c e^{-\beta_tm^{\star}}+(\es_t-\ei_t)+\frac{\maxofU}{2}\beta_t'\right]u_t\\
		&+2\maxofU\beta_t' \sqrt{u_t}\nonumber\\
		&+2(\es_t-\ei_t).\nonumber
		\end{align}
		Using that, $\forall x \in \mathbb{R},~ \frac{1}{4}x^2+1\ge x$, we get:
		\begin{align}\label{before gron}
		u_t'\le& ~2 \left[-c e^{-\beta_tm^{\star}}+(\es_t-\ei_t)+(\frac{\maxofU}{2}+\frac{\maxofU}{4})\beta_t'\right]u_t\\
		&+2\maxofU\beta_t'+2(\es_t-\ei_t).\nonumber
		\end{align}
		Let $A_t=2c e^{-\beta_tm^{\star}}	$	and $B_t=2(\es_t-\ei_t)+2\maxofU\beta_t'$.\medskip\\
		Applying Gr\"onwall's Lemma for the previous relation gives:
		\begin{equation}\label{Gron}
		u_t\le u_0 e^{\int_0^t-A_s+B_s \mathrm{d}s} +\int_0^t B_s e^{\int_s^t-A_h+B_h \mathrm{d}h} \mathrm{d}s.
		\end{equation}		
		Under Assumptions $\ref{assumption_subsec}$, there exist $b,d>0$ such that $\beta_t=b\log(1+td)$. This implies:
		$$ \beta'_t=\dfrac{bd}{1+td} \mbox{ and } e^{-m^{\star}\beta_t}=\left(\dfrac{1}{1+td}\right)^{m^{\star}b}.  $$
		Using the definition of $\es_t,\ei_t$ and the fact that $n_t=(1+t)^{\alpha}$ one can check that:
		$$ A_t=\mathcal{O}\left(\frac{1}{t^{m^{\star}b}}\right)\mbox{ and } B_t=\mathcal{O}\left(\frac{1}{t}\vee \frac{\log t}{t^{\alpha/2}}\right)$$
		We can see that if $B_t=o(A_t)$ the second term of \eqref{Gron} is bounded and gives us a finite upper bound on $u_t$. This happens as soon as: 
		\begin{equation}\label{relation1}
		m^{\star}b< 1 \wedge \alpha/2
		\end{equation}
		
		However, the condition given by \eqref{relation1} is sufficient yet not necessary. For $\alpha>2$, $B_t$ becomes of the order $\mathcal{O}(1/t)$ and thus we can choose $d$ in a way that preserves a finite upper bound of \eqref{Gron} even for $m^{\star}b=1$. One can check by direct computation that this is true for any $d<cm^{\star}/\maxofU$. 
		
		Let $\beta_t$ and $n_t$ be chosen in order to comply to one of the two previously mentioned  conditions. Then there exists a constant $K'$ such that 
		\begin{equation}\label{u is bounded}
		u_t\le K'\mbox{ for all } t\in \mathbb{R}^{+}
		\end{equation}    
		To complete the proof of  \thmref{thm:maintheorem} one can observe that for all $ t\in \mathbb{R}_+$, and all  $ \epsilon>0$ :
		$$
		\Pr(\tilde{X}_t\in \bXe)=
		\mathbb{E}(\mathds{1}_{[J^{\star}+\epsilon,+\infty)}(J(\tilde{X}_t))).$$ 
		Using the Cauchy-Schwarz inequality and the upper bound given by \eqref{u is bounded} we obtain:
		\begin{align}\label{ftbound}
		\mathbb{E}(\mathds{1}_{[J^{\star}+\epsilon,+\infty}(J(\tilde{X}_t)))
		=&\int_{\mathbb{R}}\mathds{1}_{[J^{\star}+\epsilon,+\infty}(J(x))f_t d\mu_{\beta_t}(x)\nonumber\\
		\le&\left(\int_{\mathbb{R}}(f_t)^2 d\mu_{\beta_t}(x) \right)^{\frac{1}{2}}\left(\int_{\mathbb{R}}\mathds{1}^2_{[J^{\star}+\epsilon,+\infty)}(J(x))d\mu_{\beta_t}(x)\right)^{\frac{1}{2}}\\
		\le& \|f_t\|_{L^2_{\mu_{\beta_t}}}(\mu_{\beta_t}(\bXe))^{1/2}\nonumber\\
		\le & K(\mu_{\beta_t}(\bXe))^{1/2}\nonumber
		\end{align}
		with $K=\sqrt{K'+1}$. This completes the proof of \thmref{thm:maintheorem}.
	\hfill\BlackBox\\
	\subsection{Convergence rate}
	A first rate of convergence can be deduced from \thmref{thm:maintheorem} using the concentration speed of the Gibbs measure on $\Xe$. 
	
	\begin{align}\label{concentration}
	\mu_{\beta_t}(\bXe)&= \frac{\sum_{x\in \bXe}e^{-\Bt J(x)}}{\sum_{x\in E}e^{-\Bt J(x)}}\nonumber \\
	&= \frac{\sum_{x\in \bXe}e^{-\Bt J(x)}}{\sum_{x\in \bXe}e^{-\Bt J(x)}+\sum_{x\in \Xe}e^{-\Bt J(x)}}\nonumber\\
	&\le \frac{(|E|-|\Xe|)e^{-\Bt (J^{\star}+\epsilon)}}{0+|\Xe|e^{-\beta_t J^{\star}}}\nonumber\\
	&\le \left(\frac{|E|}{|\Xe|}-1\right)(1+td)^{-b \epsilon}
	\end{align}
	As the dependency of $K$ (\thmref{thm:maintheorem}) in $b$ and $\alpha$ is not explicit, we can however not deduce an optimal choice of $(b,\alpha)$ from this bound. This can be achieved if we assume that \eqref{relation1} holds and distinguish the two cases $\alpha\le2$ and $\alpha>2$. Indeed, we can then improve the bound on $u_t$ and derive a more accurate convergence rate of the algorithm. This rate can then be optimized to obtain either an upper bound of the probability of convergence to $\Xe$ for a fixed computational budget or the minimal computational budget at a fixed risk of convergence out of $\Xe$.

	 \begin{theorem} \label{thm:conv_speed}
		Under assumptions of Section \ref{assumption_subsec}, suppose: $\beta_t=b\log(td+1)$, $n_t=(1+td)^{\alpha}$ and $m^{\star}b<\min(\alpha/2,1)$:
		\begin{itemize}
			\item if $\alpha\ge 2$, let $b$ be such that $m^{\star}b<1$ and let $\gamma\in (0,\alpha/2-m^{\star} b)$,\\
			Then, there exist $\Gamma_{\gamma},\Gamma_2>0 $ such that for t large enough, for all $\epsilon>0,$
			\begin{align*}
			\Pr(\tilde{X}_t\in \bXe) &\le  \Gamma_{\gamma}\Gamma_2(1+td)^{(m^{\star}b-1-b\epsilon)/2}+\Gamma_2(1+td)^{-b\epsilon}
			\end{align*}
			
			\item if $\alpha<2$, let $b$ be such that $m^{\star}b<\alpha/2$ and let $\gamma\in (0,\alpha/2-m^{\star} b)$,\\
			Then, there exist $\Gamma_{\gamma},\Gamma_2>0 $ such that for t large enough, for all $\epsilon>0,$
			\begin{align*}
			\Pr(\tilde{X}_t\in \bXe) &\le  \Gamma_{\gamma}\Gamma_2(1+td)^{(m^{\star}b-\alpha/2+\gamma-b\epsilon)/2}+\Gamma_2(1+td)^{-b\epsilon}
			\end{align*}\hfill\BlackBox 
		\end{itemize}
	
	\end{theorem}
	\begin{rem}
		$\gamma$ is not a new parameter of the NSA algorithm. This is a technical element that enables the tuning of the computational complexity bounds of \seqref{comp_complex}. As shown in the Appendix \ref{ann:proof_bound_ut}, $\Gamma_{\gamma}$ is of the order of $1/\gamma$. 
	\end{rem}
	\begin{rem}
	This two bounds display the trade off between the convergence rate of the Gibbs measure to the uniform distribution over the global minima and the rate of convergence of the NSA process to the Gibbs measure. For the first bound, considering $\alpha>2$ we recover the classical rate of convergence of the simulated annealing in the noise free case. This corresponds to the result of \cite{gutjahr1996simulated}. The second bound provides the rate of convergence for a choice of $\alpha<2$. It can be seen that $b$ will have to be reduced to ensure the convergence and thus this bound exhibits clearly the trade off between cooling and estimation.
	\end{rem}
		
	{\bf Proof.}
		Under assumptions of \thmref{thm:conv_speed}, the following bound \hioana{ \eqref{ut:bound2}} on $u_t$ can be derived from Gr\"onwall's Lemma (for details see Appendix \ref{ann:proof_bound_ut}):
		\begin{equation}\label{ut:bound2}
		u_t\le \begin{cases}
		\Gamma_{\gamma}(1+td)^{m^{\star}b-1} \quad  \mbox{if } \alpha\ge 2\\
		\Gamma_{\gamma}(1+td)^{m^{\star}b-\alpha/2+\gamma}\quad \mbox{if }  \alpha< 2 
		\end{cases}
		\end{equation}
		Thus we can compute a new bound on the probability that $\tilde{X}_t$ does not belong to the optimal set $\Xe$ (\textit{cf.} \ref{def:optimality_set}):
		\begin{align}\label{proba ineq}\nonumber
		\Pr(\tilde{X}_t\in \bXe)=&\int_{\mathbb{R}}\mathds{1}_{\bXe}(J(x))f_t d\mu_{\beta_t}(x)\\ \nonumber
		&=\int_{\mathbb{R}}\mathds{1}_{\bXe}(J(x))(f_t-1) d\mu_{\beta_t}(x)+\int_{\mathbb{R}_+}\mathds{1}_{\bXe}(J(x))d\mu_{\beta_t}(x)\\ \nonumber
		&\le \left(\int_{\mathbb{R}_+}(f_t-1)^2 d\mu_{\beta_t}(x) \int_{\mathbb{R}_+}\mathds{1}^2_{\bXe}(J(x))d\mu_{\beta_t}(x)\right)^{1/2}+\mu_{\beta_t}(\bXe)\\ 
		&\le \sqrt{u_t\mu_{\beta_t}(\bXe)}+\mu_{\beta_t}(\bXe)
		\end{align}
		This means that if there exist $(\alpha,b)$ such that $u_t=\mathcal{O}(\mub)$ the convergence rate in the noisy case will be of the same order as in the classical one, but for a smaller $b$.

		Using the previous inequality, \eqref{ut:bound2} and the concentration rate of the Gibbs measure given by \eqref{concentration} we have:
		\begin{align*}
		\Pr(\tilde{X}_t\in \bXe) &\le\begin{cases}
		\Gamma_{\gamma}\Gamma_2(1+td)^{\frac{m^{\star}b-1-b\epsilon}{2}}+\Gamma_2(1+td)^{-b\epsilon} \quad  \mbox{if } \alpha\ge 2\\
		\Gamma_{\gamma}\Gamma_2(1+td)^{\frac{m^{\star}b-\alpha/2+\gamma-b\epsilon}{2}}+\Gamma_2(1+td)^{-b\epsilon}\quad \mbox{if }  \alpha< 2 
		\end{cases}
		\end{align*}
		where $\Gamma_2=\frac{|E|}{|\Xe|}-1$.
	\hfill\BlackBox\\
	\subsection{Computational complexity of NSA}\label{comp_complex}
	Given the convergence rate of the algorithm, we can define  $T^{\star}$ such that the confidence inequality constraint is satisfied at time $T^{\star}$.
	
	Let $N_{call}^T$ be the number of cost function evaluations made by the NSA until time $T$. This is a random variable. We define the computational cost of the algorithm as the expectation of this random variable. It can be written as:
	$$\mathbb{E}\left(N_{call}^T\right) =\expe\left(\sum_{k\geq 1}\mathds{1}_{T_k<T}N_{T_k}\right).$$  
	 \begin{lem}\label{lemma_tstar}
		Let $\delta,\epsilon>0$, $\gamma\in (0,\alpha/2-m^{\star} b)$ and $$T^{\star}=\frac{1}{d}\left(\max\left( \left(\frac{2\Gamma_{\gamma}}{\delta}\right)^{2/(\cv -m^{\star}b+b\epsilon)},\left(\frac{2\Gamma_2}{\delta}\right)^{1/b\epsilon}\right)-1\right).$$ Then, for all $t\ge T^{\star}$, $\Pr(\tilde{X}_t\in \bXe)\le \delta$ and the computational cost up to time $T^{\star}$ is bounded: 
		$$\mathbb{E}\left(N_{call}^{T*}\right)\le \frac{1}{d}\max\left( \left(\frac{2\Gamma_{\gamma}}{\delta}\right)^{2(\alpha+1)/(\cv -m^{\star}b+b\epsilon)},
		\left(\frac{2\Gamma_2}{\delta}\right)^{(\alpha+1)/b\epsilon}\right).$$ 
		
	\hfill\BlackBox
\end{lem}
	{\bf Proof.}
		In order to prove this statement, we use the inequalities from \thmref{thm:conv_speed} treating each term separately.  \\
		We consider $T_1,T_2$ such that $\Gamma_{\gamma}\Gamma_2(1+dT_1)^{m^{\star}b-\cv-b\epsilon}=\delta/2$ and $\Gamma_2(1+dT_2)^{-b\epsilon}=\delta/2$. This implies:
		
		\begin{equation}\label{definition of T}
		1+dT_1= \left(\frac{2\Gamma_{\gamma}}{\delta}\right)^{2/(\cv -m^{\star}b+b\epsilon)} \mbox{ and }
		1+dT_2=\left(\frac{2\Gamma_2}{\delta}\right)^{1/b\epsilon}.
		\end{equation}
		Now we can define $T^\star$, the time after which the current state of the NSA belongs to $\Xe$ with probability at least $1-\delta$, \textit{i.e.}, $\forall t>T^\star,~\Pr(\tilde{X}_t\in \Xe)\ge 1-\delta$: $$T^{\star}=\max(T_1,T_2).$$ 
		
		We are interested in the computational cost up to time $T^{\star}$, more precisely the expected  number of Monte Carlo simulations used up to $T^{\star}$.  This is given by $\expe\left(\sum_{k\geq 1}\mathds{1}_{T_k<T^{\star}}N_{T_k}\right)$. The value of this quantity cannot be computed exactly, but it can easily be upper bounded.
			\begin{align*}\nonumber
		\expe\left(\sum_{k\le 1}\mathds{1}_{T_k<T^{\star}}N_{T_k}\right)=&\expe\left(\expe\left(\sum_{k\ge1}\mathds{1}_{T_k<T^{\star}}N_{T_k}\right)|(T_k)_{k=1\dots+\infty}\right)\\ \nonumber
		=&\expe\left(\sum_{k\ge1}\mathds{1}_{T_k<T^{\star}}n_{T_k}\right)\\ 
		\le&\expe\left(\sum_{k\ge1}\mathds{1}_{T_k<T^{\star}}\right)n_{T^{\star}}
		\end{align*}
		The last inequality is implied by the fact that $n_t$ is an increasing function. Since $\sum_{k\ge1}\mathds{1}_{T_k<T^{\star}}$ is a Poisson variable of parameter $T^{\star}$, using the definition of $n_t$ one can see that: 
		\begin{align}\label{cost ineq}
		\expe\left(\sum_{k\le 1}\mathds{1}_{T_k<T^{\star}}N_{T_k}\right)\le&T^{\star}(1+dT^{\star})^{\alpha}\\
\nonumber		\le&\frac{1}{d}(1+dT^{\star})^{\alpha+1}.
		\end{align}
		We conclude using \eqref{definition of T}.
	\hfill\BlackBox\\

	The rate of growth of the total computation number is mainly driven by the exponent of $\frac{1}{\delta}$ in the cost function. We are looking for the couple $(\alpha,b)$ that minimizes this quantity and fulfills the requirements of \thmref{thm:conv_speed}. We can split the problem into two sub-problems:
	\begin{enumerate}
		\item[Case 1:]$\ad-\gamma> 1$
		\begin{align}
		&\underset{b,\alpha}{\min}~\max\left(\frac{2(\alpha+1)}{1-m^{\star}b+b\epsilon},\frac{\alpha+1}{b\epsilon}\right)\label{optNSApbmfacile}\\
		s.t.&\nonumber\\
		& m^{\star}b<1 \mbox{ and } \alpha-2\gamma> 2 \nonumber
		\end{align}
		\item[Case 2:]$\ad\le1$
		\begin{align}
		&\underset{b,\alpha}{\min}~\max\left(\frac{2(\alpha+1)}{\alpha/2-\gamma-m^{\star}b+b\epsilon},\frac{\alpha+1}{b\epsilon}\right)\label{optNSApbm}\\
		s.t.&\nonumber\\
		& 0<\gamma<\ad -m^{\star}b \mbox{ and } \alpha-2\gamma\le 2\nonumber
		\end{align} 
	\end{enumerate}

	The solution of \eqref{optNSApbmfacile} is obvious, the minimal value for $\alpha$ and the maximal for $b$, \textit{i.e}, $\alpha$ must be as close to $2$ as possible  and $b=\frac{1}{m^{\star}+\epsilon} $.
	As for \eqref{optNSApbm}, we consider two sub-cases. First suppose that: 
	\begin{equation}\label{first case}
	\frac{2(\alpha+1)}{\alpha/2-\gamma-m^{\star}b+b\epsilon}\geq\frac{\alpha+1}{b\epsilon} \Longleftrightarrow \alpha/2-\gamma-m^{\star}b\le b\epsilon.
	\end{equation} 
	The function we want to minimize is strictly decreasing in $\alpha$ and strictly increasing in $b$, so its minimum value is attained for the maximal value of $\alpha$ and the minimal value of $b$, under the domain constraints given by \eqref{optNSApbm} and \eqref{first case}, so the solution is: 
	\begin{equation}\label{solution}
	\alpha=2(1+\gamma) \mbox{ and } b>\frac{1}{m^{\star}+\epsilon}.
	\end{equation}
	In the second sub-case, supposing that the inequality \eqref{first case} is inverted, the problem can be resumed at minimizing $ (\alpha+1)/b\epsilon$, a decreasing function  with respect to $b$, for 
	$$b\le \frac{\alpha/2-\gamma}{(m^{\star}+\epsilon)} \mbox{ and } \alpha\le 2.$$
	Replacing $b$ by its maximal value the objective function becomes a decreasing function in $\alpha$, and therefore we obtain the same solution as before, defined in \eqref{solution}. 
	This is a quite comprehensive result, as it indicates that the lower the required accuracy in the solution space is, \textit{i.e.}, $\epsilon$ increases and thus the size of $\Xe$ does too, the faster the temperature can decrease to zero. We need to explore less the state space. 
	
	\begin{coro}\label{corollary}
		For the optimal parameters choice defined in \eqref{solution}, an $\epsilon$-optimal solution is returned by NSA  with probability $1-\delta$ at a computational cost at most : 
		$$\frac{1}{d}\left(\frac{2\Gamma_{\gamma}}{\delta}\right)^{\frac{m^{\star}+\epsilon}{\epsilon}(3+2\gamma)},$$
		where $\Gamma_{\gamma}$ is defined in \thmref {thm:conv_speed}.  
	\end{coro}
This is rather costly but represents a general bound with few constraints on the function $J$. However, if the function $J$ has additional properties the bound can be significantly improved: 

\begin{coro}\label{corollary2}
	Suppose that $J$ has no well containing a local minimum, apart from the one containing the global minimum, \textit{i.e.} $m^{\star}=0$, then  an $\epsilon$-optimal solution is returned by NSA  with probability $1-\delta$ at a computational cost at most : 
\begin{align*}
\left(\dfrac{2\log \frac{1}{\delta}}{d\epsilon}\right)^{3}.
\end{align*}
\end{coro}
\begin{rem}
	We recover the polynomial dependency in $1/\epsilon$ and $\log 1/\delta$ of the state-of-the-art complexity results (\textit{c.f.} \cite{woodroofe1972normal} and \cite{nemirovski1982problem}) which are of the order of $\epsilon^{-2}\log(\frac{1}{\delta})$ for strongly convex cost functions. As we relax this assumption and only consider cost functions with no local minimum, it seems coherent to observe a slight degradation of the complexity. 
\end{rem}
{\bf Proof.} In order to have an estimate of the computational cost in this setting we follow the same method as before and highlight only the main steps of the proof. 
First remark that in this case, Theorem \ref{thHolley} states that there exist $C,c>0$ such that $\forall \beta \in \mathbb{R}^+$:
\begin{equation}\label{spectral gap m zero}
c\le \gamma(\beta) \le C
\end{equation}
 This changes the differential inequality obtained for $u_t=\| f_t-1\|_{\mub}^2$ and thus \eqref{before gron} becomes :
\begin{align}
		u_t'\le& ~2 \left[-c +(\es_t-\ei_t)+(\frac{\maxofU}{2}+\frac{\maxofU}{4})\beta_t'\right]u_t+2\maxofU\beta_t'+2(\es_t-\ei_t).\nonumber
		\end{align}
		We can apply Gr\"onwall's Lemma and obtain the same type of inequality as before: 
		\begin{equation}\label{Gron m zero}
		u_t\le u_0 e^{\int_0^t-A_s+B_s \mathrm{d}s} +\int_0^t B_s e^{\int_s^t-A_h+B_h \mathrm{d}h} \mathrm{d}s.
		\end{equation}		
		where $B_t$ has the same form as before, $ B_t=2(\es_t-\ei_t)+2\maxofU\beta_t'$ and $A_t=2c$ . 	

The convergence of $u_t$ towards $0$ can be proved now for a larger class of functions $n_t,\beta_t$, since: $$ A_t=\mathcal{O}(1) \mbox{ and }B_t=\mathcal{O}(\beta_t/\sqrt{n_t}\vee \beta_t').$$	
We no longer need to impose $\beta_t=\mathcal{O}(\log_t )$. Let $\alpha,b,d>0$.   Define $$n_t=(1+t)^{\alpha} \mbox{ and } \beta_t=d(1+t)^{b}.$$
	Using \eqref{Gron m zero} one can check that we have a finite upper bound on $u_t$ as soon as: 
	$$\{ b<\alpha/2\wedge 1, d>0\} \mbox{ or } \{b=1, \alpha\ge 2, 0<d<c\}. $$
	This in particular implies that the NSA algorithm converges a.s. to the set of global minimums of $J$. Furthermore for the first set of conditions one can prove using the same technique as in Appendix \ref{ann:proof_bound_ut} that :
	$$u_t=\mathcal{O}(t^{\max(b-1,b-\alpha/2)}).  $$
	This means that there exits $\Gamma_{\gamma}'>0$ such that for $t$ large enough $u_t\le \Gamma_{\gamma}' t^{-\gamma}$, where $\gamma=-\max(b-1,b-\alpha/2)$. Using this, \eqref{proba ineq} and \eqref{concentration}, for $t$ large enough, we get:
\begin{align*}
\Pr(\tilde{X}_t\in \bXe)&\le \Gamma_{\gamma}' t^{-\gamma/2}e^{-\epsilon d(1+t)^b/2}+\Gamma_2 e^{-\epsilon d(1+t)^b}\\
&\le e^{-\epsilon d(1+t)^b/2}(\Gamma_{\gamma}' t^{-\gamma/2}+\Gamma_2 e^{-\epsilon d(1+t)^b/2})\\
&\le e^{-\epsilon d(1+t)^b/2}
\end{align*}
The last inequality is valid as soon as $t>\max\left[\left(\frac{2\log(2\Gamma_2)}{\epsilon d}\right)^{\frac{1}{b}}-1, \left(2\Gamma_{\gamma}'\right)^{\frac{2}{\gamma}}\right]$. This is not a restrictive condition. Take for example the minimization of the $\|.\|_1$ over the subset subset  $E=\{x\in\mathbb{Z}^p, \|x\|_{\infty}\le n\}$ for some $n\in \mathbb{N}$. As $\Gamma_2=\|E\|-1=(n+1)^p-1$, the time for which the first part of the condition is fulfilled only grows linearly with the dimension of the search space. We show latter on that the optimal choice for $b$ is one and thus the second part of the condition can be omitted.

 Let $\delta>0$ be a fixed. Using the previous inequality one can compute $T^{\star}$ such that the confidence inequality constraint is satisfied:
 $$T^{\star}_{\epsilon,\delta}=\left(-\dfrac{2 \log \delta}{d\epsilon }\right)^{1/b}-1.$$ 
 Regarding the computational cost we remind the reader that \eqref{cost ineq} implies:
 \begin{align*}
 \expe (N^{T^{\star}}_{call})&\le n_{T^{\star}} T^{\star}\le \left(-\dfrac{2\log \delta}{d\epsilon}\right)^{\frac{1+\alpha}{b}}.
 \end{align*}
  We can optimize this bound with respect to $\alpha$ and $b$ in the same way as for \corref{corollary}. This leads to  $\alpha=2$ and $b<1$ and thus to the desired results:  
  \begin{align*}
  \expe (N^{T^{\star}}_{call})\le \left(-\dfrac{2\log \delta}{d\epsilon}\right)^{3}.
  \end{align*}
  	\hfill\BlackBox
	\section{Numerical experiments}\label{sec:num_res}
	In this section we first present some test cases, for which we use an additive Gaussian noise at each evaluation. We recover the theoretical results introduced by \cite{gutjahr1996simulated}. In a second part we present some results for the aircraft trajectory optimization problem. In this case the solution of the problem is unknown. We can only observe the total cost improvement in comparison with a trajectory optimized for a similar but deterministic setting. 
	
\begin{figure} 
		\centering
		\includegraphics[width=0.8\linewidth]{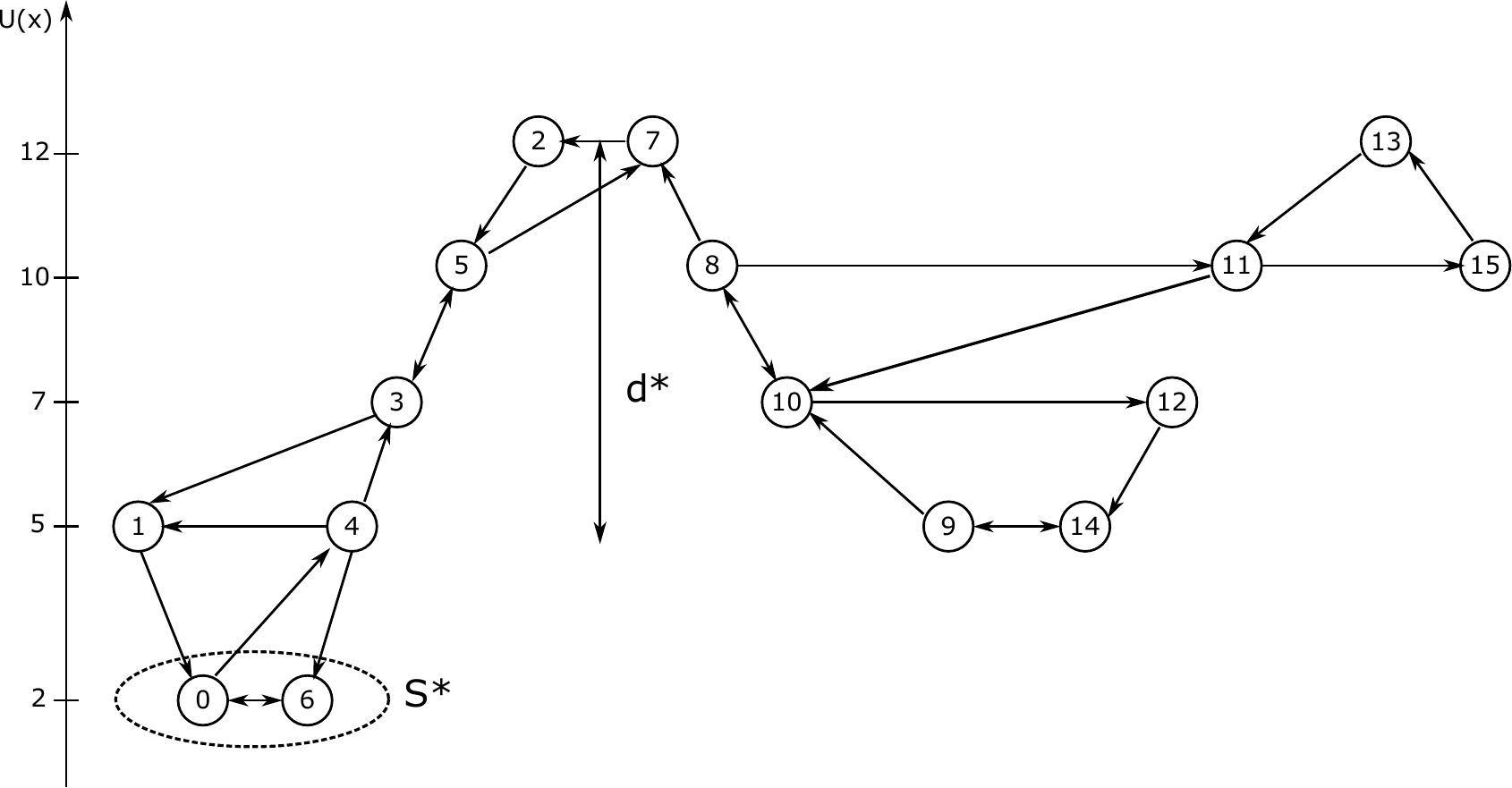}
		\caption{B. Hajek test case for the simulated annealing in a deterministic environment}
		\label{fig:Hajek}
	\end{figure}

	\paragraph{Basic exemple}~\\
	The first experimental setting we consider, was introduced in \cite{hajek1988cooling}. The cost function and the neighbourhood structure are represented on Figure \ref{fig:Hajek}.  
	This is of particular interest as the function has two basins from which it is hard to escape. B. Hajek has shown that the following holds:
	
	\noindent
	\begin{theorem}[\cite{hajek1988cooling}]~\\
		If $\beta_k=b~log(k+2)$, then \quad$b\le d^{\star}\Leftrightarrow \lim\limits_{k\to\infty}\Pr(X_k\in S^{\star})=1,$ 	where $d^{\star}$ is the maximum depth of a cup containing a local but not global minima. The depth of a cup is the maximal energy difference between two of it states  and $(X_k)_{k\in \mathbb{N}}$ denotes the Markov chain generated by the classical simulated annealing. \hfill\BlackBox
	\end{theorem}
	
 For a complete definition of $d^{\star}$, see \cite{hajek1988cooling}.
	\begin{figure} 
		\centering
		\includegraphics[width=0.7\linewidth]{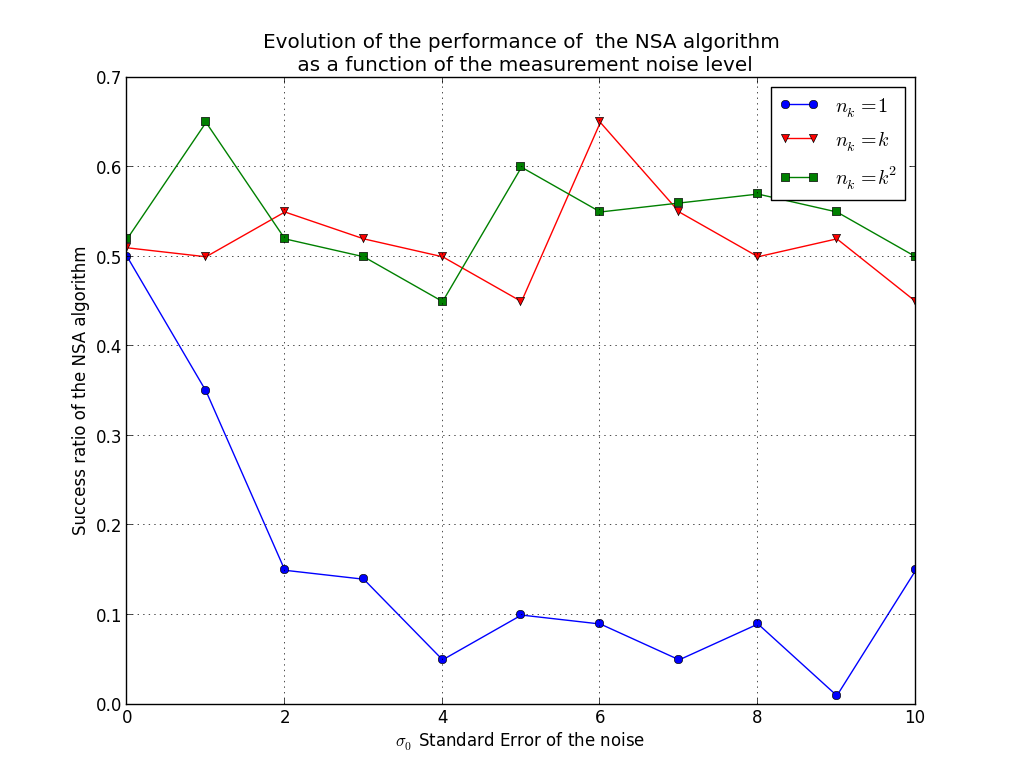}
		\caption{Convergence performance of NSA for the Hajek setting}
		\label{fig:simu_classiques}
	\end{figure}
	
\begin{figure} 
		\includegraphics[width=\linewidth]{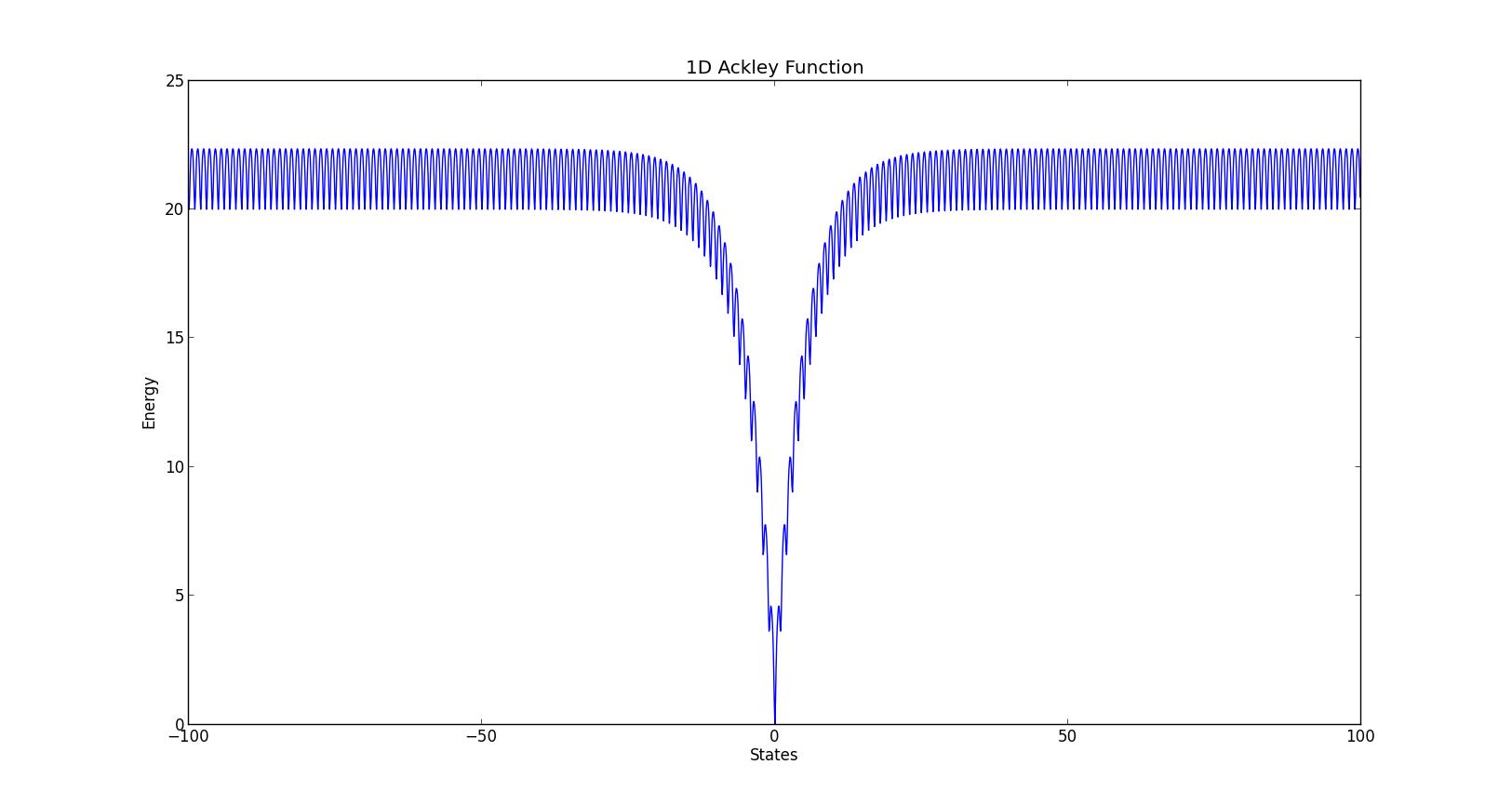}
		\caption{Ackley 1D Test Function }
		\label{fig:ackley1d}
	\end{figure}

\begin{figure} 
		\centering
		\includegraphics[width=0.7\linewidth]{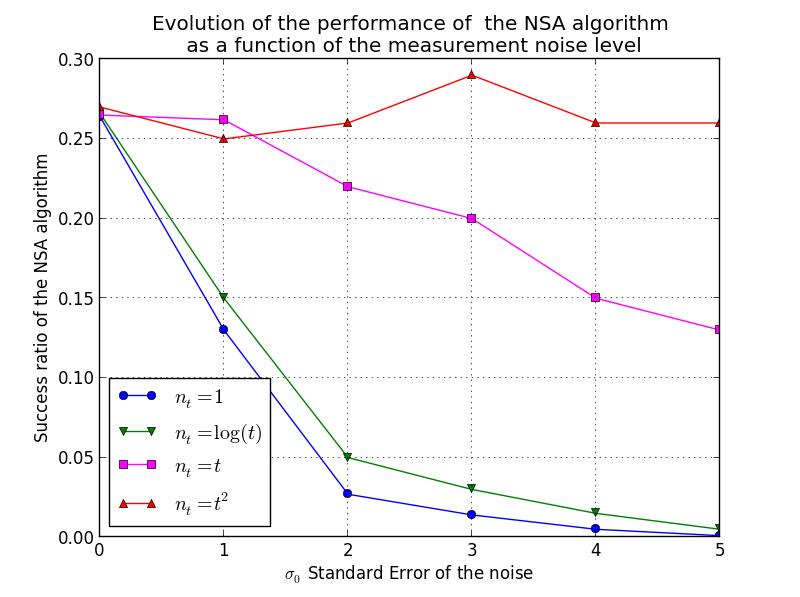}
		\caption{Performance of the NSA algorithm as a function of the level of noise on the evaluation of the cost function}
		\label{fig:NSA_perfo}
	\end{figure}

	We add Gaussian noises to the cost function of Figure \ref{fig:Hajek} with different variance levels to highlight the fact that if no sampling is performed the simulated annealing performance becomes rapidly very poor as the variance increases. On the other hand it appears that the performance of the NSA for a linear increase of the mean number of samples is as good a quadratic one. These results are summarized on Figure \ref{fig:simu_classiques}.
	\paragraph{Ackley test function}~\\
	We introduce a second test case to further asses these observations. We consider the uniformly (2000 points) discretized version of the Ackley function in one dimension on $[-100,100]$. This function has many local minima as shown on Figure \ref{fig:ackley1d}.
	Figures \ref{fig:NSA_perfo} displays the convergence results for different levels of variance of the noise for each estimation schedule introduced in this paper. We observe that the only case where the convergence is not impacted by the noise variance increase is the $n_t=t^2$ case. 
	
\begin{figure} 
		\centering
		\includegraphics[width=0.7\linewidth]{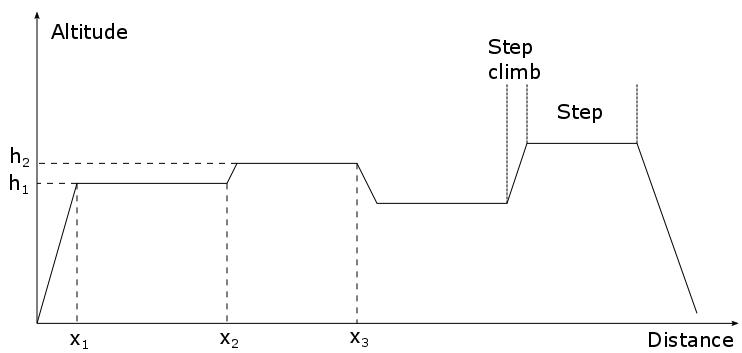}
		\caption{Aircraft trajectory, structure of the vertical path}
		\label{fig:traj_theorique}
	\end{figure}

\begin{figure} 
		\centering
		\includegraphics[width=0.7\linewidth]{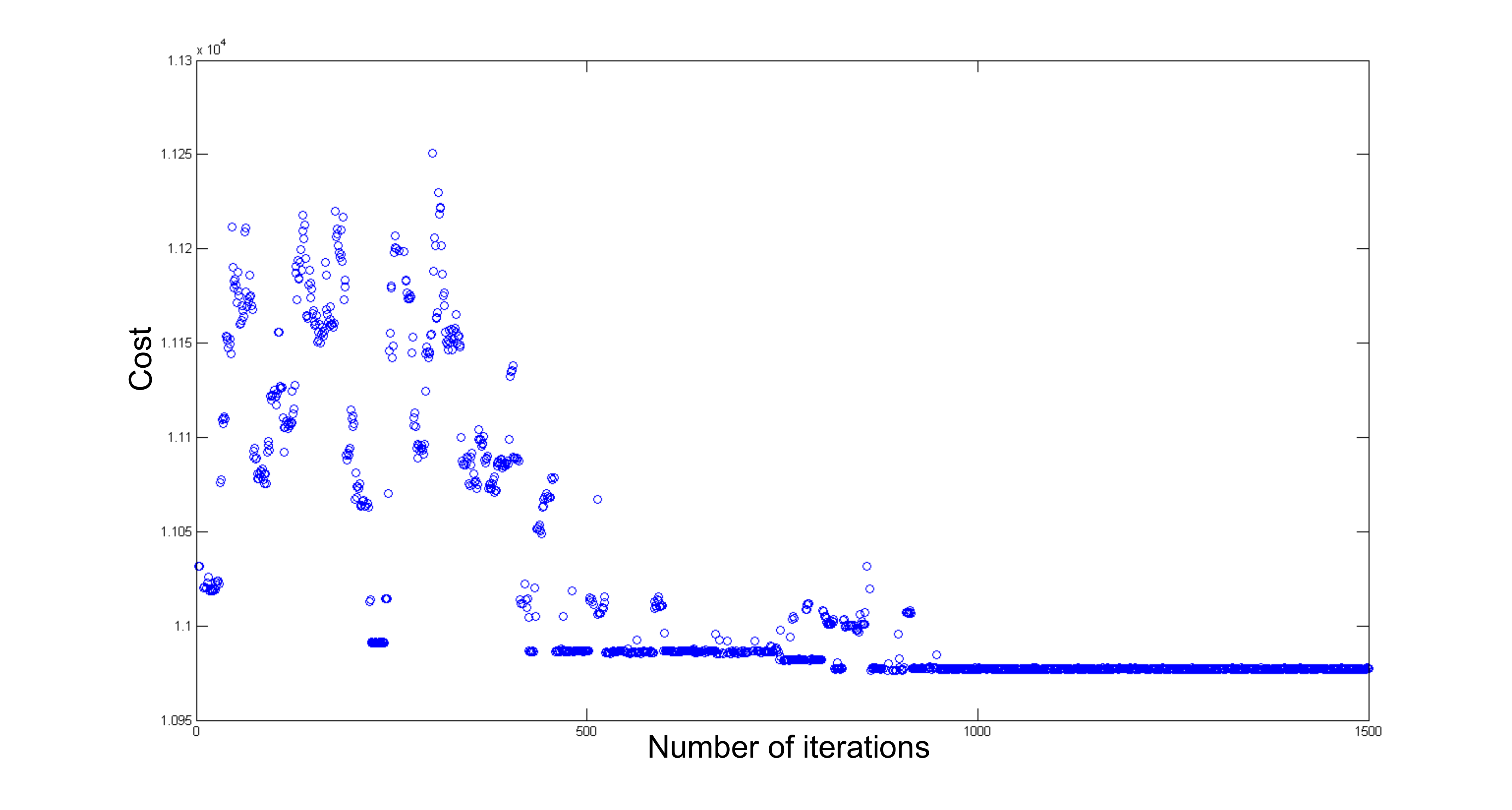}
		\caption{NSA descent: Aircraft trajectory optimization problem}
		\label{fig:NSA2}
	\end{figure}

	These results highlight the fact that a logarithmic sampling schedule is not appropriate in general, even in the Gaussian case. This invalidates partially the hypotheses introduced in \cite{fink1998inverse}. A clear gap is highlighted between the linear and the quadratic schedule.\\
	\paragraph{Aircraft trajectory optimization}~\\
	 We use a black box trajectory evaluator for a long range commercial aircraft. We consider a direct shooting method for optimizing the vertical part of the trajectory. As displayed on Figure \ref{fig:traj_theorique}, the vertical path is made of a sequence of flight segments at constant altitude called steps. The transitions between those steps are called step climbs. This has been put in place by the international authorities to ease the air traffic control. Aircraft can only fly at a finite set of altitudes. The steps climbs are transition phases that must be very short. The Figure \ref{fig:traj_theorique} is a conceptual. It does not reflect the real scale of the different phases. Our optimization variables are the vectors of position of the steps and the vector of steps' altitude, denoted respectively $x$ and $h$ on Figure \ref{fig:traj_theorique}. The structure of this airspace strongly limits the number of steps. We will only consider the problem with an \textit{a priori} number of steps. 
	There are two main reasons why the aircraft might vary its altitude during a flight (optimizing fuel consumption and air traffic control). Because of fuel consumption, the aircraft weight is decreasing during the flight. Analyzing the laws of flight physics, it can be shown that there exists an altitude at which the fuel consumption per flown distance unit is minimal. It can also be shown that this altitude increases as the weight decreases. This last statement is however only true if there is no wind. It is easily understandable that for some particular wind map configuration it might be preferable to target lower altitudes at lower weights.\\
	The choice of the vertical path must be declared to the authorities before the flight to ensure traffic manageability.  Airlines operating aircraft have therefore a stochastic optimization problem to solve. This is a stochastic problem for two main reasons. First, they only access predicted weather conditions that suffer some uncertainty. Second, the airspace is not empty and sometimes air traffic controllers might refuse some altitude changes because of the presence of other aircraft. As the weather, the traffic is not known in advance.\\
	We applied NSA to the problem of finding an optimal 3 steps configuration. An example of the current solution cost evolution with respect to the number of iterations is displayed on Figure \ref{fig:NSA2}.  
	
\begin{figure}
		\centering
		\includegraphics[width=0.7\linewidth]{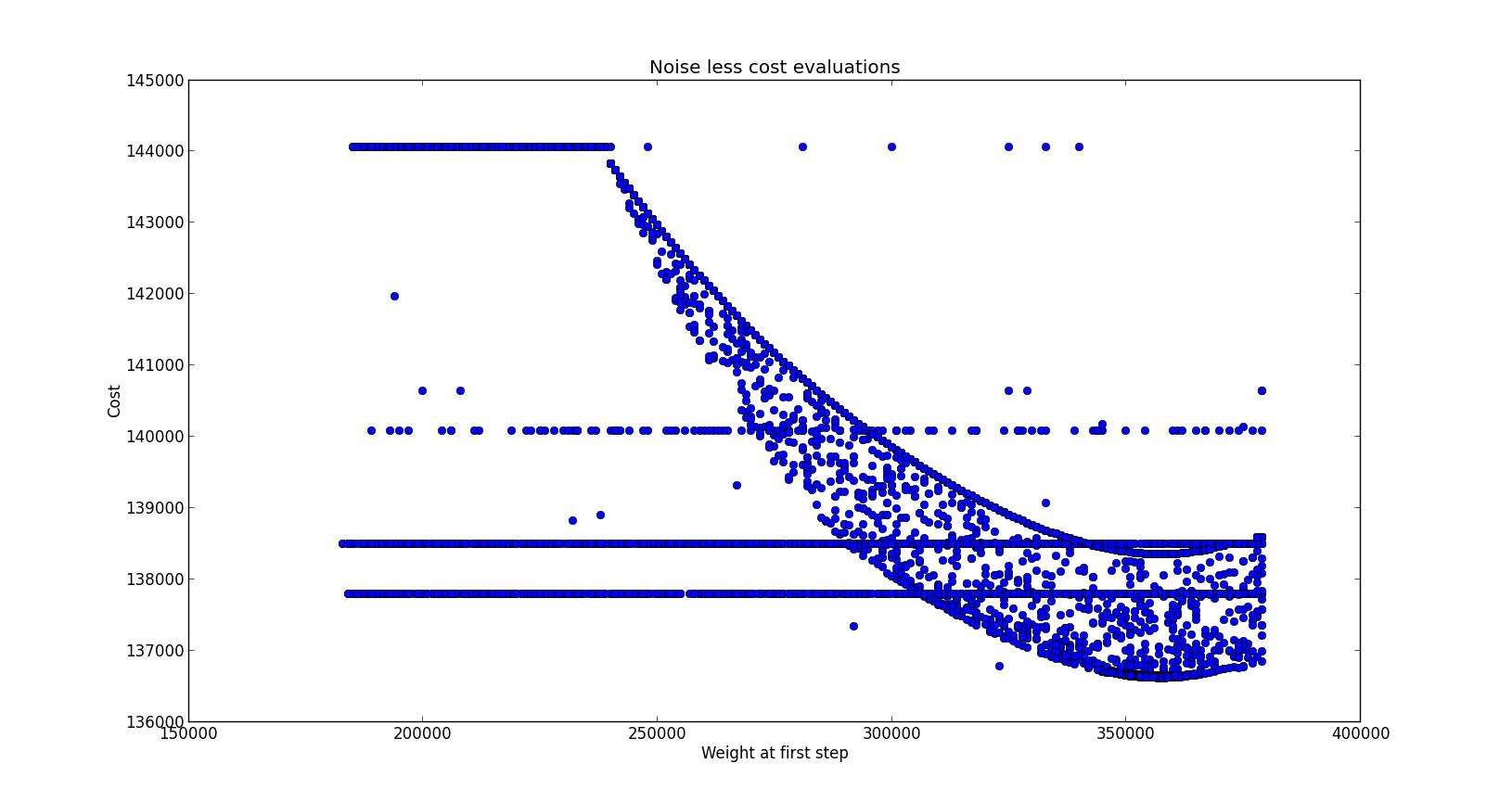}
		\caption{Sampling of the cost function along the first step position for a 3 step vertical path }
		\label{fig:cost_expe}
	\end{figure}

	We observe a very quick convergence to a low cost trajectory. We do not claim it is a general behaviour. It might be due to the structure of the cost function. Figure \ref{fig:cost_expe}, shows how the cost evolves with respect to the ground position of the first step. It is obviously not convex but has some regularity. We can observe some flat parts. This explains why gradient based methods would fail solving this problem. 
	
	As for the previous problems we have observed that the increase sampling condition must be satisfied to ensure a good behaviour of the algorithm.

	\appendix
	
	\section{Proof of bound \eqref{ut:bound2}}\label{ann:proof_bound_ut}
	{\bf Proof.}[Proof of \thmref{thm:conv_speed}]~\\
	Let $\beta_t=b\log(td+1)$ and	 $n_t=(1+td)^{\alpha}.$\\ 
	Recall \eqref{gronineq}:
	\begin{align*}
	u_t'\le& ~2 \left[-c e^{-\beta_tm^{\star}}+(\epsilon^+-\epsilon^-)+\left(\frac{\maxofU}{2}+\frac{\maxofU}{4}\right)\beta_t'\right]u_t +2\maxofU\beta_t'+2(\epsilon^+-\epsilon^-)\nonumber
	\end{align*}
	Let $A_t=2c e^{-\beta_tm^{\star}}	$	and $B_t=2(\epsilon^+-\epsilon^-)+2\maxofU\beta_t'$.\medskip\\
	Applying Gr\"onwall's Lemma for the previous relation gives:
	\begin{equation*}
	u_t\le u_0 e^{\int_0^t-A_s+B_s \mathrm{d}s} +\int_0^t B_s e^{\int_s^t-A_h+B_h \mathrm{d}h} \mathrm{d}s
	\end{equation*}		
	Under Assumptions $\ref{assumption_subsec}$, there exist $b,d>0$ such that $\beta_t=b\log(1+td)$. This implies:
	$$ \beta'_t=\dfrac{bd}{1+td} \mbox{ and } e^{-m^{\star}\beta_t}=\left(\dfrac{1}{1+td}\right)^{m^{\star}b}.  $$
	Using the definition of $\ei, \es$ we have:
	$$\es-\ei_t=2\sqrt{2\pi}\beta_t\sigma\left[\frac{e^{\frac{\beta_t^2\sigma^2}{2(1-\delta)n_t}}}{\sqrt{(1-\delta)n_t}}+e^{\beta_t^2\sigma^2-an_t}\right].$$
	This implies that when $t$ goes to infinity: 
	$$ A_t=\mathcal{O}\left(\frac{1}{t^{m^{\star}b}}\right)\mbox{ and } B_t=\mathcal{O}\left(\frac{1}{t}\vee \frac{\log t}{t^{\alpha/2}}\right)$$
	In order to highlight the mains ideas of the proof we will try to simplify the notations as much as possible. First observe that for all $\alpha>0$ and all $0<\gamma<\alpha/2$ , $\dfrac{\log t}{t^{\alpha/2}}=o\left(\dfrac{1}{t^{\alpha/2-\gamma}} \right)$.\\

	Hence we can assume there exist $A,B>0$ and $\cx, \cva$ such that:
	$$ A_t= A\frac{d}{(1+td)^{\cx}}\mbox{ and } B_t\le B\frac{d}{(1+td)^{ \cva}}$$
	where $\cx=m^{\star}b$ and $ \cva= \min(1,-\gamma+ \alpha/2)$. Since $\min(1,\alpha/2)>m^{\star}b$, and $\gamma$ can be chosen arbitrarily  close to $0$, we choose it such that $\cx< \cva$. This means that $0<\gamma<\alpha/2-m^{\star}b$. \\
	\begin{rem}
		The choice of $\gamma$ influences the choice of $B$. If $B_t=\mathcal{O}\left(\dfrac{\log t}{t^{\alpha/2}}\right)$, there exists $C_B>0$ such that $B_t\le C_B \dfrac{\log t}{t^{\alpha/2}}, \forall t$. The constant $B$ is then such that  $\forall t$, $C_B\dfrac{\log t }{t^{\gamma}}\le B$. Hence we can choose:
		\begin{equation}\label{size of gamma}
		B=\dfrac{C_B}{e\gamma}.
		\end{equation}	  
	\end{rem}	
	Let $T_t^1=u_0 e^{\int_0^t-A_s+B_s \mathrm{d}s}$ and  $T_t^2=\int_0^t B_s e^{\int_s^t-A_h+B_h \mathrm{d}h} \mathrm{d}s$.\\

	The first term $T_t^1$ is always easy to deal with and one can check that under the theorem's assumptions we always have $T_t^1= o(1/t^{\cx- \cva})$ when $t$ goes to infinity. As for the second term, using a substitution gives:
	\begin{align}\label{ineqAB}
	T_t^2&\le\int_{1}^{1+td} \frac{B}{s^{ \cva}} e^{\int_s^{1+td}-\frac{A}{h^{\cx}}+\frac{B}{h^{ \cva}} \mathrm{d}h} \mathrm{d}s \nonumber \\
	&\le e^{-\frac{A(1+td)^{1-\cx}}{1-\cx}+\frac{B(1+td)^{1- \cva}}{1- \cva}}\int_{1}^{1+td} \frac{B}{s^{ \cva}} e^{\frac{As^{1-\cx}}{1-\cx}-\frac{Bs^{1- \cva}}{1- \cva} } \mathrm{d}s
	\end{align}
	For the last inequality we assume $ \cva\neq1$ which corresponds to the case $\alpha \le 2$.
	
	Let $I_t=\int_{1}^{1+td} \frac{B}{s^{ \cva}} e^{\frac{As^{1-\cx}}{1-\cx}-\frac{Bs^{1- \cva}}{1- \cva} } \mathrm{d}s$ and $f_s= \frac{As^{1-\cx}}{1-\cx}-\frac{Bs^{1- \cva}}{1- \cva}$. Let $T_0$ be such that for all $s\ge T_0$,  $s^{ \cva-\cx}\ge  \frac{B+1}{A}$ (for instance $T_0=(\frac{B}{A}+1)^{1/(\cva-\cx)}$).  We can write $I_t$ as follows:
	\begin{align*}
	I_t&=\int_{1}^{T_0} \frac{B}{s^{ \cva}} e^{f_s } \mathrm{d}s+\int_{T_0}^{1+td} \frac{B}{s^{ \cva}} e^{f_s } \mathrm{d}s\\
	&=K_{T_0}+\int_{T_0}^{1+td}\dfrac{B}{s^{ \cva}(As^{-\cx}-Bs^{- \cva})} e^{f_s}f'_s \mathrm{d}s\\
	&=K_{T_0}+\left[ \dfrac{B}{As^{ \cva-\cx}-B} e^{f_s}\right]_{T_0}^{1+td}+\int_{T_0}^{1+td}\dfrac{AB( \cva-\cx)s^{ \cva-\cx-1}}{(As^{ \cva-\cx}-B)^2} e^{f_s} \mathrm{d}s\\
	\end{align*}
	Since $\cx< \cva$, $\dfrac{A( \cva-\cx)s^{-\cx-1}}{(As^{ \cva-\cx}-B)^2}$ goes to $0$ when $s$ goes to infinity. Moreover one can check that for all  $s\ge T_0$ this quantity is smaller than $1/2$. Using this we get:
	$$I_t\le K_{T_0}+\left[ \dfrac{B}{As^{ \cva-\cx}-B} e^{f_s}\right]_{T_0}^{1+td}+\dfrac{1}{2}I_t\\
	$$		  
	and therefore for all $t\ge T_0$:
	$$I_t\le 2\left[ \dfrac{B}{As^{ \cva-\cx}-B} e^{f_s}\right]_1^{1+td}+2K_{T_0}\\
	$$		  
	This gives:	
	\begin{align*}
	T_t^2&\le 2e^{-f_{1+td}}\left(\left[ \dfrac{B}{As^{ \cva-\cx}-B} e^{f_s}\right]_1^{1+td}+K_{T_0}\right)\\
	&\le 2\left[ \dfrac{B}{A(1+td)^{ \cva-\cx}-B} \right]+2\left(K_{T_0}-\dfrac{B}{A-B}e^{f_1}\right)e^{-f_{1+td}}\\
	\end{align*}	
	Regrouping the terms we obtain for all $t\ge T_0$:
	\begin{equation}\label{ineq u}
	u_t\le \left[u_0+2\left(K_{T_0}-\dfrac{B}{A-B}e^{f_1}\right)\right] e^{-f_{1+td}}+ \dfrac{2B}{A(1+td)^{ \cva-\cx}-B}
	\end{equation}
	Since the first term is a $\mathcal{O}(e^{-f_t})$ and therefore a $o(t^{\cx- \cva})$ it is obvious that:
	\begin{equation}\label{order of u}
	u_t=\mathcal{O}\left( \dfrac{1}{t^{ \cva-\cx}}\right)\mbox{ when } t\to \infty
	\end{equation}
	This means that for all $\gamma\in (0,\alpha/2-m^{\star}b)$ there exists $\Gamma_{\gamma}>0$ such that:
	
	\begin{equation}\label{u gamma}
	u_t\le \Gamma_{\gamma} t^{-\alpha/2+m^{\star}b+\gamma} \mbox{ for all } t\ge T_0.
	\end{equation}
	Since  $f_t=\mathcal{O}(t^{1-\cx})$,  $e^{-f_t} t^{\cx- \cva}$ goes very fast to zero and therefore the size of $\Gamma_{\gamma}$ is mainly driven by $ \dfrac{2B}{A}$. Using \eqref{size of gamma} one can see that:$$\Gamma_{\gamma}\simeq \frac{1}{\gamma}.$$
	
	If $\alpha>2$,  $ \cva=1$ and \eqref{ineqAB} becomes of the form:
	$$T_t^2\le e^{-\frac{A(1+td)^{1-\cx}}{1-\cx}+B\log(1+td)}\int_{1}^{1+td} \frac{B}{s^{1+B}} e^{\frac{As^{1-\cx}}{1-\cx}}\mathrm{d}s.$$
	\begin{rem}
		In this case one can choose $\gamma=\alpha/2-1$, this way $\cv=1$ and $\Gamma_{\gamma}$ is minimal. 
	\end{rem}
	
	\hfill\BlackBox\\
	
\section{Definition of $m^\star$}\label{mstardefequiv}
	In this section we prove that the definition of $m^\star$, \textit{i.e.} \eqref{definition of m}, is equivalent to the definition provided by \cite{Holley88Simu}.
	\begin{lem}
			Let, $m^{\star}_{HS}:=\underset{x,y \in E}{\max}\left\{\underset{p \in P_{xy}}{\min} \left\{\underset{z\in p}{\max} ~ J(z)   \right\}- J(y)-J(x)+\min_{u}J(u) \right\}$.\\
			Then 
			$$m^{\star}=m^{\star}_{HS}$$
	\end{lem}
	\hfill\BlackBox\\
		{\bf Proof.} Let $x,y\in E$ and denote $H_{xy}:=\underset{p \in P_{xy}}{\min} \left\{\underset{z\in p}{\max} ~ J(z)   \right\}$.\\
	First it can be noticed that if $x$ is a global minimum of $J$ then we have
	\begin{align}
	H_{x,y}- J(y)-J(x)+\min_{u}J(u)=H_{x,y}- J(y)
	\end{align}
	Thus  $m^{\star}_{HS}\ge H_{x^{\star},y}- J(y)$ for any $y$ in $E$, where $x^{\star}$ is a global minimum of $J$.\\ 
	
	Recall that $m^{\star}=\underset{x,y \in E}{\max}\left\{H_{xy}- \max\left(J(y),J(x)\right)\right\}$.
	As the set of paths going from $x$ to $y$ containing a global minimum $x^{\star}$ is a subset of the paths going from $x$ to $y$, we have:
	$$ H_{xy}\le \max\left(H_{x^{\star}x},H_{x^{\star}y}\right)$$
	Let $x,y\in E$ such that $m^{\star}=H_{xy}- \max\left(J(y),J(x)\right)$,
	\begin{align*}
	m^{\star}&\le \max\left(H_{x^{\star}x},H_{x^{\star}y}\right)-\max\left(J(y),J(x)\right)\\
	&\le \max\left(H_{x^{\star}x}-J(x),H_{x^{\star}y}-J(y)\right)\\
	&\le m^{\star}_{HS}
	\end{align*}
	On the other hand, as $\pourtout{x,y}{E}$  we have $-\min\left(J(y),J(x)\right)+\min_{u}J(u)\le 0$, so
	\begin{align*}
	 H_{xy}- J(y)-J(x)+\min_{u}J(u)\le H_{xy}- \max\left(J(y),J(x)\right)
	\end{align*}
	This implies $m^{\star}_{HS}\le m^{\star}$, which completes the proof.
	\hfill\BlackBox\\

	\bibliographystyle{plain}
	\bibliography{biblio}
\end{document}